\documentclass[10pt,twocolumn,letterpaper]{article}

\usepackage{cvpr}

\usepackage{titletoc}

\definecolor{cvprblue}{rgb}{0.21,0.49,0.74}
\usepackage[pagebackref,breaklinks,colorlinks,allcolors=cvprblue]{hyperref}

\usepackage{booktabs}
\usepackage{multirow}
\usepackage{makecell}
\usepackage{graphicx}
\usepackage[table]{xcolor}
\usepackage{tikz}

\definecolor{colyellow}{HTML}{FFD56B}
\definecolor{colgray}{HTML}{7F7F7F}
\definecolor{colorange}{HTML}{FFA06B}
\definecolor{colgreen}{HTML}{4ECDC4}

\newcommand{\circnumwhite}[2]{
\tikz[baseline=(char.base)]{

  \node[
    shape=circle,
    fill=#1,
    text=white,
    font=\bfseries,
    inner sep=0.03em,
    outer sep=0pt,
    minimum size=0.1em,
    align=center
  ] (char) {#2};
}
}

\usepackage[most]{tcolorbox}
\usepackage{xcolor}

\definecolor{SCOUT1}{RGB}{90,120,255}
\definecolor{SCOUT2}{RGB}{120,110,240}
\definecolor{SCOUT3}{RGB}{150,100,225}
\definecolor{SCOUT4}{RGB}{180,90,210}
\definecolor{SCOUT5}{RGB}{210,80,195}

\definecolor{SCOUT1}{RGB}{64,180,255}
\definecolor{SCOUT2}{RGB}{72,200,240}
\definecolor{SCOUT3}{RGB}{80,220,225}
\definecolor{SCOUT4}{RGB}{88,235,210}
\definecolor{SCOUT5}{RGB}{96,250,195}

\definecolor{SCOUT1}{RGB}{255,120,80}
\definecolor{SCOUT2}{RGB}{255,100,100}
\definecolor{SCOUT3}{RGB}{235,80,120}
\definecolor{SCOUT4}{RGB}{215,60,140}
\definecolor{SCOUT5}{RGB}{195,40,160}

\definecolor{mapcolorRGB}{RGB}{101,114,156}
\colorlet{mapcolor}{mapcolorRGB}

\definecolor{langcolorRGB}{RGB}{181,133,142}
\colorlet{langcolor}{langcolorRGB}

\definecolor{colabcolorRGB}{RGB}{112,141,129}
\colorlet{colabcolor}{colabcolorRGB}

\newtcbox{\cmdtag}{enhanced,nobeforeafter,tcbox raise base,boxrule=0.4pt,top=0mm,bottom=0mm,
  right=0mm,left=4mm,arc=2pt,boxsep=2pt,before upper={\vphantom{dlg}},
colframe=mapcolor!50!black,coltext=mapcolor!25!black,colback=mapcolor!10!white,
    overlay={\begin{tcbclipinterior}\fill[mapcolor!80] (frame.south west)
        rectangle node[text=white,font=\sffamily\bfseries\tiny,rotate=90] {CMD} ([xshift=4mm]frame.north west);\end{tcbclipinterior}}}

\newcommand{\mytitle}{\textcolor{SCOUT1}{A}\textcolor{SCOUT2}{U}\textcolor{SCOUT3}{R}\textcolor{SCOUT4}{A}
}
\newcommand{\Namesystemtwo}{multi-modal encoder\xspace}
\newcommand{\ModelName}{AURA\xspace}
\newcommand{\CmdEncoder}{Spatial-Aware Instruction Encoder\xspace}

\newcommand{\CmdEncodershort}{SIE\xspace}
\newcommand{\automode}{Autopilot\xspace}
\newcommand{\takeovermode}{Takeover\xspace}

\definecolor{lightorange}{RGB}{255, 218, 185}
\definecolor{OursGold}{RGB}{250,245,220}
\newcommand{\ours}[0]{\rowcolor{OursGold}}

\def\paperID{1726}
\def\confName{CVPR}
\def\confYear{2026}

\title{\mytitle: Multimodal Shared Autonomy for Real-World Urban Navigation}

\author{
    Yukai Ma$^{1,2}$
    \quad
    Honglin He$^{1}$
    \quad
    Selina Song$^{1}$
    \quad
    Wayne Wu$^{1}$
    \quad
    Bolei Zhou$^{1}$ \\
    $^{1}$ University of California, Los Angeles
    \quad
    $^{2}$ Zhejiang University \\
    \textbf{\url{https://vail-ucla.github.io/aura/}}
}

\begin{document}

\maketitle

\begin{abstract}
Long-horizon navigation in complex urban environments relies heavily on continuous human operation, which leads to fatigue, reduced efficiency, and safety concerns. Shared autonomy, where a Vision-Language AI agent and a human operator collaborate on maneuvering the mobile machine, presents a promising solution to address these issues. However, existing shared autonomy methods often require humans and AI to operate within the same action space, leading to high cognitive overhead. We present Assistive Urban Robot Autonomy (\ModelName), a new multi-modal framework that decomposes urban navigation into high-level human instruction and low-level AI control. \ModelName incorporates a {\CmdEncoder} to align various human instructions with visual and spatial context. To facilitate training, we construct \textit{MM-CoS}, a large-scale dataset comprising teleoperation and vision-language descriptions. Experiments in simulation and the real world demonstrate that {\ModelName} effectively follows human instructions, reduces manual operation effort, and improves navigation stability, while enabling online adaptation.
Moreover, under similar takeover conditions, our shared autonomy framework reduces the frequency of takeovers by more than 44\%. Demo video and more detail are provided in the project page.
\vspace{-15pt}
\end{abstract}
    
\section{Introduction}
\label{sec:introduction}

\begin{figure}[t!]
    \centering
    \includegraphics[width=1\linewidth]{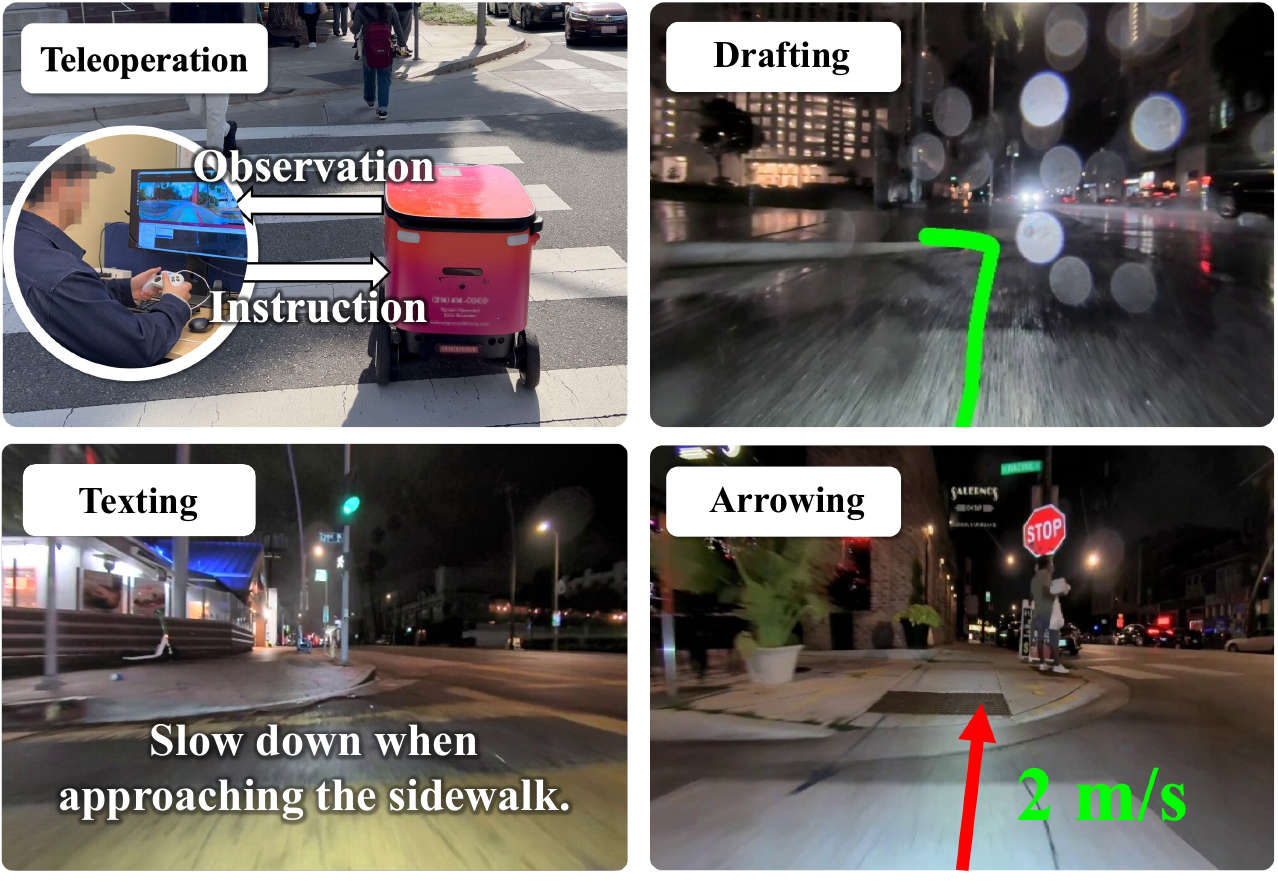}
    % \vspace{-0.25in}
    \caption{
    \textbf{Shared Autonomy for Urban Navigation.} We introduce {\ModelName}, a dual-system VLA for shared autonomy in urban navigation. {\ModelName} not only follows instructions but also enables human users to guide and correct a robot in real time through various visual and language instructions.
    }
    % \vspace{-0.225in}
    \vspace{-10pt}
    \label{fig:teaser}
\end{figure}
Despite rapid progress in autonomous driving on roads, AI remains significantly challenging to safely operate mobile machines in public urban spaces, such as sidewalks, parks, and school campuses, due to the complexity of the surroundings and the diverse range of human activities. Thus, many existing sidewalk mobile machines (e.g., assistive wheelchairs and food delivery bots) still rely heavily on a human-in-the-loop approach. For example, remote workers may teleoperate or closely supervise delivery bots to accomplish last-mile tasks. In these settings, the human pilot must remain fully attentive to both control and situational awareness. Road hazards, such as rugged terrain and fragmented curbs, as well as unexpected pedestrian behavior and other human-made errors, pose significant risks to mobile machines and their human pilots.

To address these challenges, recent studies have explored the paradigm of shared autonomy, where AI agents assist and augment human pilots in maneuvering machines in both training~\cite{cai2025predictive, cai2025robot, peng2023learning} and test-time phases~\cite{udupa2023shared}. For long-horizon navigation tasks, researchers are increasingly seeking AI assistant models that can effectively reduce human workload, allowing humans to remain in the loop primarily for monitoring and failure recovery~\cite{sridhar2024nomad,liu2024citywalker,shah2023vint,hirose2025learning}.

Deploying navigation autonomy models in the real world for long-horizon navigation typically involves switching between full human operation and full AI control~\cite{cai2025predictive, cai2025robot, peng2023learning, hirose2025learning, sridhar2024nomad, liu2024citywalker}. However, these approaches assume that the human and AI operate in the same low-level action space that directly controls the driving wheels and pedals, thereby requiring the human to operate at the same frequency as the AI. This coupling becomes inefficient and cognitively demanding for long-horizon tasks such as urban sidewalk food delivery.
We believe a more effective paradigm is a \textit{division of labor} between humans and AI by decomposing urban navigation into hierarchical levels: (i) humans provide high-level instructions, such as reasoning about corner cases and proposing alternative routes, and (ii) AI agents make low-level executions, such as lane keeping and obstacle avoidance. By delegating low-level control to the AI while reserving high-level strategic decisions for the human, this human-AI shared autonomy framework will achieve improved efficiency, safety, and robustness in long-horizon urban navigation.

In this work, we propose \textbf{A}ssistive \textbf{U}rban \textbf{R}obot \textbf{A}utonomy (\textbf{\ModelName}), a multi-modal shared autonomy framework designed to understand human instructions and handle low-level control, thereby significantly reducing human operation costs. Specifically, {\ModelName} is a Vision-Language-Action (VLA) model with a dual-system architecture. Acting as an ``autonomous assistant,'' {\ModelName} can be seamlessly integrated into existing delivery robots without any hardware modifications.
For high-level instruction, the system enables various human instructions through a vision-and-language interface, where concise natural-language or visual prompts on live video streams replace laborious joystick control. It enables safe, scalable sidewalk autonomy while substantially reducing operator workload.
As illustrated in the center of Fig.~\ref{fig:teaser}, {\ModelName} can interpret multiple forms of human instructions: (i) texting, where users describe the intent; (ii) drafting, where users draw a rough path on the observation view; and (iii) arrowing, where users demonstrate the desired speed and direction. 

A key challenge in shared autonomy is interpreting ambiguous human instructions and grounding them within the surrounding spatial context. To address this, we introduce a Spatial-Aware Instruction Encoder ({\CmdEncodershort}) that explicitly aligns textual instructions with both the semantic layout and the geometric structure of the scene. This design addresses the spatial-understanding limitations of standard vision–language models (VLMs), enabling the model to reason about the user's intent and where those instructions can be executed, thereby improving robustness across diverse real-world environments.

To support this model training, we construct a multimodal video dataset, MM-CoS, on top of our sidewalk teleoperation dataset CoS~\cite{he2026learning} that includes a total of 50 hours of high-quality teleoperation data collected on real-world sidewalks. The dataset spans a diverse range of scenarios, including different cities, weather conditions, and lighting variations. We further curate relevant clips using behavior- and ego-state-based filters to ensure that the resulting human control behaviors are representative. By combining human instructions inferred from VLMs with ground-truth trajectories from human operators, we obtain a unified dataset in which multimodal human instructions serve as inputs and expert trajectories serve as outputs.

Experiments show that our framework accurately interprets and executes human instructions, achieving over 15\% lower L2 error and reducing human operation costs by more than 70\% compared to baselines. These results highlight the effectiveness of our shared-autonomy design in enabling efficient and human-aligned navigation.
In summary, our contributions are threefold:
\begin{itemize}
\item A multi-modal shared autonomy framework that integrates human instruction following with low-level control via a unified VLM encoder and diffusion policy.
\item We introduce {\CmdEncoder} designed for instruction understanding, which is trained on a new dataset with multi-modal instructions to connect human teleoperation and VLM-based intention inference.

\item Simulated and real-world experiments demonstrate the effectiveness of our approach in following instructions, improving stability, and reducing human operation cost.
\end{itemize}

\section{Related Work}
\noindent\textbf{Human-in-the-loop Learning and Shared Autonomy.} 
It is crucial to ensure safety when deploying robots in urban environments, and incorporating human preferences into decision-making offers a promising way toward trustworthy, human-centered autonomy~\cite{peng2023learning, cai2025predictive}. Prior works have incorporated human feedback during training to improve policy alignment or to correct undesired behaviors, a paradigm often referred to as learning with human involvement or human-in-the-loop learning~\cite{zhang2017query, abel2017agent, wang2021appli, xu2022look, peng2023learning, cai2025predictive, peng2025data}. In this paradigm, humans actively intervene or provide demonstrations when the robot exhibits unsafe or suboptimal behaviors, allowing the policy to learn corrective actions and refine itself through shared autonomy. During deployment, it is equally important for the robot to understand implicit and explicit human intentions and preferences, expressed through multimodal instructions such as language descriptions, visual cues, etc., as demonstrated in recent advances~\cite{wang2021appli, xu2022look, shi2024yell, lukitchenvla, cui2023no, karamcheti2022shared, aronson2024intentional, ji2025pref}. However, language-based guidance is typically limited to high-level task specifications and cannot support safety-critical, high-frequency interactions required for applications such as urban navigation. In these scenarios, the policy must be able to rapidly adjust its behavior in response to human instructions such as visual paths or corrective joystick inputs, enabling fine-grained shared autonomy that complements language.

\noindent\textbf{Urban Navigation.} Navigation is an important problem in robot learning, requiring autonomous agents to perceive, plan, and act safely in complex real-world environments. It poses challenges for understanding human behavior and ensuring safety and compliance in densely populated urban spaces. Early approaches focused primarily on map-based navigation~\cite{moravec1985high, thrun2002probabilistic, durrant2006simultaneous}, relying on accurate maps and localization techniques such as SLAM~\cite{mur2015orb} to estimate position and plan collision-free paths. However, these methods often assume static environments and struggle to adapt to dynamic, socially interactive settings like urban sidewalks. Recent works~\cite{chen2017socially, fan2020distributed, mirowski2016learning} based on reinforcement learning (RL)~\cite{sutton1998reinforcement, lillicrap2015continuous} have demonstrated promising results in mapless navigation by eliminating the dependency on maps. However, these methods often have limited generalizability, particularly in visual navigation~\cite{shen2019situational,putta2024embodiment,truong2021learning}, due to the simulation-to-real gap~\cite{tobin2017domain, peng2018sim, yu2024learning}. Inspired by the success of scaling laws in language modeling~\cite{kaplan2020scaling, achiam2023gpt}, many recent works have proposed various vision-based navigation foundation models~\cite{shah2023gnm, shah2023vint, sridhar2024nomad, liu2024citywalker, hirose2024lelan, bar2024navigation}, leveraging massive video data for improved generalizability across different robot platforms and camera configurations.

\noindent\textbf{Robotic Instruction Following.}
Advances in large language models (LLMs)~\cite{kaplan2020scaling,achiam2023gpt} and instruction alignment~\cite{ouyang2022training} suggest that neural networks can align with human preferences. Inspired by this, embodied AI asks whether similar principles let robots follow instructions and generalize across tasks. In perception, vision-language models (VLMs)~\cite{alayrac2022flamingo,bai2023qwen,mei2024continuously,ma2025leapvad} show strong multimodal reasoning, motivating VLA policies that integrate perception, language, and low-level control~\cite{ahn2022can,black2024pi_0,intelligence2025pi_,kim2024openvla,cheng2024navila}, enabling instruction following across platforms~\cite{black2024pi_0,intelligence2025pi_,hirose2025omnivla}.

However, language instructions alone encode high-level intent and are insufficient for the fine-grained, high-frequency corrections required in dynamic navigation. The bottleneck arises from how instruction alignment is performed. Following InstructGPT and RLHF~\cite{ouyang2022training}, most VLA models rely on offline datasets. Such alignment contrasts with navigation, where safety and robustness often depend on \textit{real-time} human feedback. While recent systems explore interactive manipulation with visual prompting~\cite{xing2023dual,jiang2023vima}, these approaches primarily target short-horizon interactions and discrete action spaces, making them fundamentally different from the continuous, long-horizon trajectory reasoning required by navigation. In this work, we propose a novel shared-autonomy framework that addresses these limitations by integrating high-level human intent with real-time low-level control during inference.

\section{{\ModelName} Framework}
\label{sec:method}

% merget a and c
\begin{figure*}[t!]
    \centering
    \includegraphics[width=1\linewidth]{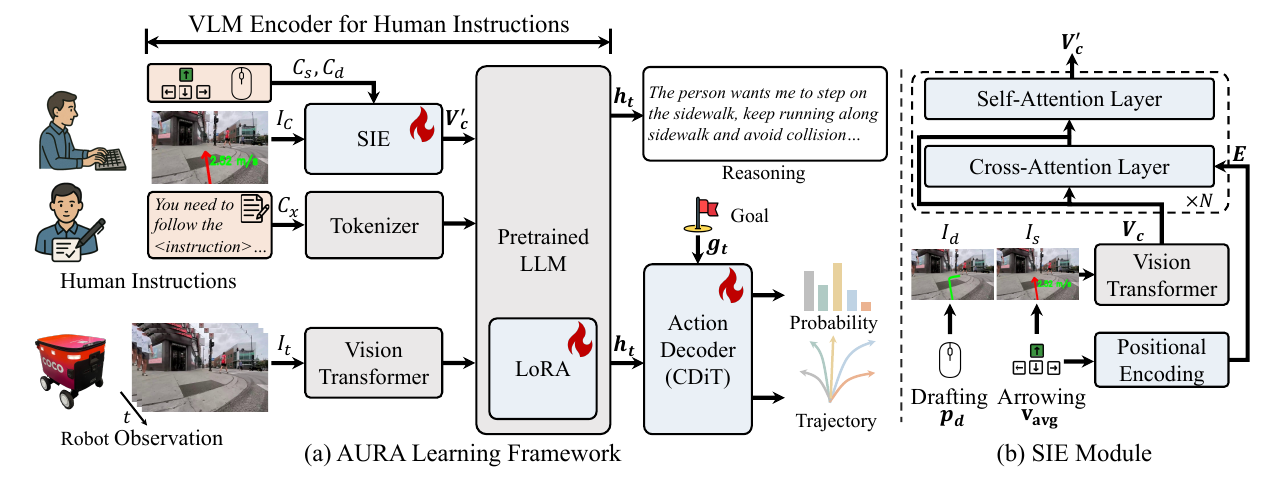}
    \vspace{-0.25in}
    \caption{
    \textbf{Overview of the AURA shared autonomy framework.}
    (a) {\ModelName} takes front-camera RGB observations and optional human guidance (e.g., texting, drafting, or arrowing). Observations are encoded by a ViT, while human inputs are processed by the SIE and tokenized; all tokens are fused in a pretrained LLM (with LoRA adapters) to produce context features. A diffusion-based action decoder then predicts a distribution over future trajectories via anchor proposals.
    (b) The SIE converts drafting/arrowing inputs into instruction tokens: it renders the human input as visual prompts, encodes the control points/vectors, and fuses them with visual features to produce instruction embeddings that are injected into the LLM via $\langle\texttt{instruction}\rangle$.
}
    \vspace{-15pt}
    \label{fig:model}
\end{figure*}

We present {\ModelName}, an end-to-end shared autonomy framework for long-horizon navigation. {\ModelName} takes diverse human instructions as input and predicts future waypoints to control mobile machines. Figure~\ref{fig:model} provides an overview of the framework and its key components.
As follows, we first formulate the shared autonomy problem in urban navigation and specify the input and output representations.
Then, we detail the model architecture in Section~\ref{sec:share autonomy framework} and introduce the key component, the {\CmdEncodershort} in Section~\ref{sec:instruction Encoder}.
Next,  we describe the dataset and annotation process to support training in Section~\ref{sec:Dataset Collection and Annotation}.
Finally, we give the model training strategies in Section~\ref{sec:Training strategy}.

\paragraph{Problem Formulation.}
\label{sec:Problem Formulation}
We aim to design and train a shared-autonomy system for mapless goal-directed visual navigation that supports various types of human instructions across diverse urban environments. The system relies solely on egocentric RGB images to perceive its surroundings.
{\ModelName} provides two navigation modes:
(i) \texttt{\automode}: {\ModelName} takes sparse GPS waypoints as input and supports basic capabilities such as sidewalk following and obstacle avoidance.
ii) \texttt{\takeovermode}: When GPS signals are unreliable, goal locations are ambiguous, or autopilot encounters corner cases it cannot handle, a human can intervene by providing guidance via texting instructions, drafting future paths, or arrowing demonstrations to prompt {\ModelName}.
This design eliminates the need for pre-built maps or explicit localization modules and frames navigation as a sequential decision-making problem.

At each timestep $t$, the agent receives a history of the past 3 frames of RGB observations $\boldsymbol{I}_{t}$. In {\automode}, a sub-goal or route $\boldsymbol{g}_t$ in egocentric coordinates is also provided. In {\takeovermode}, the sub-goal is replaced by a human instruction $C_{t'}$, provided at time $t'$, where the instruction type 
$c \in \{C_x, C_d, C_s\}$, with $t - t' \le t^{*}$, and $t^{*}$ denotes the maximum duration for which a human instruction can provide guidance. The agent $\mathcal{M}_{\theta}$ takes these inputs and outputs the action $\boldsymbol{a}_t$ to control the robot.

\subsection{Model Architecture}
\label{sec:share autonomy framework}

We describe the dual-system architecture of {\ModelName}, as illustrated in Figure~\ref{fig:model}. 
{\ModelName} comprises two main components: a {\Namesystemtwo} that encodes observations and multi-modal instructions, and a diffusion-based policy executor. Specifically, {\ModelName} leverages anchor-based regression and classification to learn future trajectory generation. A diffusion transformer (DiT) encodes the robot’s sensor configuration and target point, which are then cross-attended with high-level instruction features from the VLM backbone to produce denoised motor actions. We provide a detailed description of each module below.

\vspace{-5mm}
\paragraph{DiT Action Decoder.}
\label{sec:framework diffusion}
{\ModelName} first generates robot actions in a purely autonomous setting using a diffusion-based policy. 
Given input context features, the DiT decoder produces multi-modal future trajectories conditioned on the robot’s observations and navigation goal. 
Instead of starting from Gaussian noise, we initialize the diffusion process from $m{=}64$ trajectory anchors representing motion primitives (e.g., straight, turn, stop) clustered from the \textit{MM-CoS} dataset. 
Building on our prior work, MIMIC~\cite{he2026learning}, a lightweight transformer decoder conditions the denoising process on context features $\boldsymbol{h}_t$, navigation goal $\boldsymbol{g}_t$, and diffusion-timestep embeddings $t_d$, producing refined trajectories along with their confidence scores. 

\vspace{-5mm}
\paragraph{VLM for Human Instructions.}
\label{sec:frame vlm}
To incorporate human guidance, we augment the base policy with high-level instructions using the InternVL3-2B~\cite{zhu2025internvl3} VLM backbone. 
Past and current RGB images are resized to $448\times448$, processed by the visual encoder, and projected via an MLP to produce 256 image token embeddings per frame. 
We retain a special $\langle\text{image}\rangle$ token to represent each image context in the textual input.

Human instructions are injected via an additional $\langle\texttt{instruction}\rangle$ token, whose embeddings are produced by our {\CmdEncodershort} (see Section~\ref{sec:instruction Encoder}).
The resulting \textit{vision-language-instruction} embeddings $\boldsymbol{h}_t$ combine both image and instruction features.
During inference and policy training, we extract intermediate representations from the 12\textsuperscript{th} layer, balancing inference speed and representational quality.
In addition to providing features for control, we attach a lightweight text head to decode interpretable reasoning traces for language supervision.

Finally, these embeddings are cross-attended by the DiT action decoder, conditioning continuous trajectory generation on both robot observations and human instructions, enabling safe and flexible shared autonomy.

% ----------------------------------
\subsection{{\CmdEncoder}}
\label{sec:instruction Encoder}
{\CmdEncodershort} is a key component of the VLM encoder (Figure~\ref{fig:model}(b)), designed to embed the spatial information in human instructions.
For the \texttt{\takeovermode} mode, human instructions $C \in \{C_x, C_d, C_s\}$ provide explicit guidance through visual overlays. This requires the model to infer the human’s instruction in an immediate prompt and generate sequential actions over a period of time. To do so, the model must understand both the semantic and geometric information in the human instruction. Fortunately, ViT already possesses strong semantic understanding, which improves prompts.

Specifically, we render the instructions (e.g., trajectory lines or steering arrows) on the observation and encode the instruction image $I_{C}\in\left\{I_d,I_s\right\}$ with the same vision encoder to obtain instruction visual features $\boldsymbol{V}_c \in \mathbb{R}^{N_v \times d_v}$.

However, VLM is not inherently sensitive to spatial information. To effectively ground the instructions in their geometric context, we introduce modality-specific embeddings.
For the drafting trajectory instruction, we sample $K$ pixel coordinates $\boldsymbol{p}_d = \{(u_i, v_i)\}_{i=1}^K$ along the projected trajectory line in normalized image space. Following the design of Segment Anything~\cite{kirillov2023segment}, we apply Fourier-based positional encoding with learnable frequency basis:
\begin{equation}
\text{PE}(\boldsymbol{p}_{d,i}) = [\sin(\boldsymbol{w}^\top \boldsymbol{p}_{d,i}), \cos(\boldsymbol{w}^\top \boldsymbol{p}_{d,i})]
\end{equation}
where $\boldsymbol{w} \in \mathbb{R}^{2 \times d_p}$ is a learnable Gaussian random matrix. To preserve point ordering along the trajectory, we add learnable index embeddings: $\boldsymbol{E}_{d}^{(i)} = \text{PE}(\boldsymbol{p}_{d,i}) + \text{PosEmbed}(i)$. All point embeddings are concatenated and processed through MLP and self-attention to obtain the final encoding $\boldsymbol{E}_{d} \in \mathbb{R}^{d_v}$.

For arrowing instruction $ \mathbf{v}_{\text{avg}} = (v, \theta)$ where $v$ is speed and $\theta$ is heading angle, we use a rotation-invariant encoding that handles both forward and backward motion:
\begin{equation}
\boldsymbol{E}_s = \text{MLP}([\cos(\theta'), \sin(\theta'), \log(1 + |v|)])
\end{equation}
where $\theta' = \theta + \pi \cdot \mathbb{1}_{v < 0}$ adjusts heading for backward motion. This representation is processed through MLP and self-attention for the final embedding $\boldsymbol{E}_s \in \mathbb{R}^{d_v}$.

The geometric embeddings $\boldsymbol{E} \in \{\boldsymbol{E}_d, \boldsymbol{E}_s\}$ are fused with instruction visual features $\boldsymbol{V}_c$ via cross-attention with a residual connection, then refined through a 4-head self-attention layer and a final MLP, yielding instruction-aware features $\boldsymbol{V}'_c$ that are injected into the language model via special $\langle\texttt{instruction}\rangle$ tokens.

\subsection{Auto-Labeling for Human Instruction}
\label{sec:Dataset Collection and Annotation}
To enable the training of the AURA framework, we require data that provides high-quality explanations for actions as text instructions $C_{\text{text}}$, as well as precise odometry and camera parameters for generating visual instructions $C_{d}$ and $C_{s}$. The corresponding sequential trajectories serve as outputs in urban navigation scenarios. 

However, constructing such a dataset presents two main challenges. 
(i) \textbf{Lack of urban sidewalk data:} Prior datasets are primarily collected in campuses, indoor environments, or plazas, leaving a gap for real-world deployment. To mitigate this, we repurpose the large-scale real-world sidewalk teleoperation logs from our prior sidewalk-autopilot study~\cite{he2026learning}, which already cover diverse urban sidewalks and long-horizon navigation behaviors.
(ii) \textbf{Annotation quality:} Existing urban datasets often lack high-quality textual explanations that justify actions. We address this by generating action-grounded language explanations for each trajectory using a VLM-based captioning pipeline (Appendix~\ref{appendix:label pipeline}), producing higher-quality and more consistent textual supervision.

Recent advances in VLMs~\cite{bai2025qwen2,zhu2025internvl3} and large-scale urban navigation datasets~\cite{hirose2025learning, karnan2022socially, akhtyamov2025egowalk, shah2021rapid} have made it increasingly feasible to train such models effectively. 
To further address these limitations, we collect a diverse 50-hour teleoperation dataset comprising 3,040 trajectories captured by a wheeled robot across various real-world urban environments. 
Finally, we augment our collected data with RECON~\cite{shah2021rapid}, SCAND~\cite{karnan2022socially}, and EgoWalk~\cite{akhtyamov2025egowalk} to construct our training dataset, \textit{MM-CoS}. 

\begin{figure}[t!]
    \centering
    \includegraphics[width=1.02\linewidth]{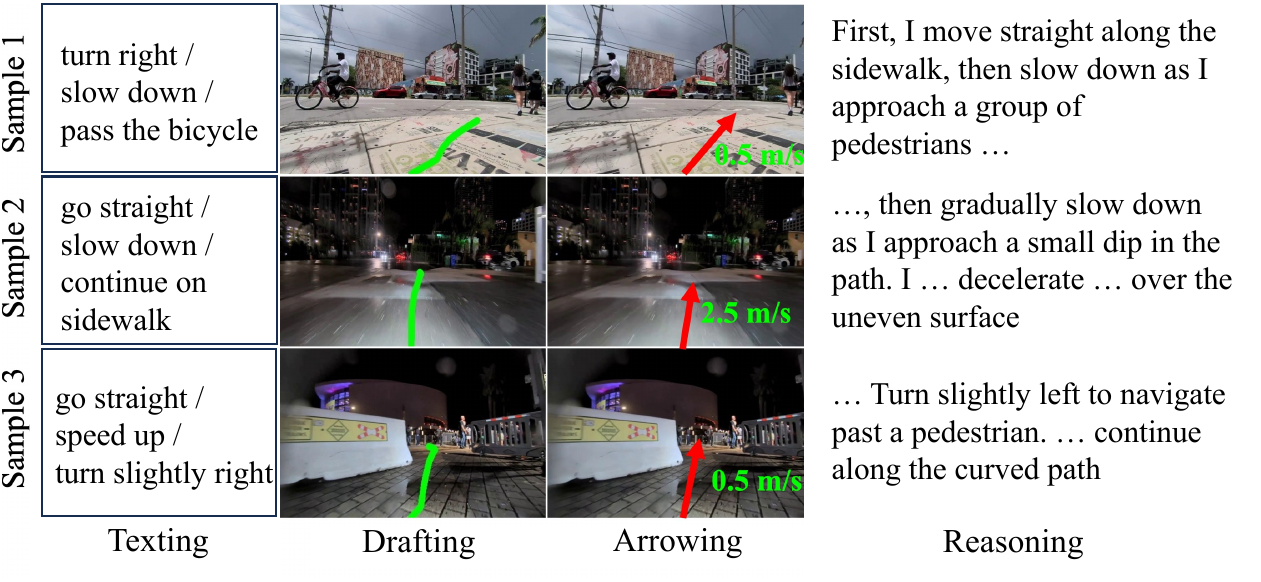}
    % \vspace{-0.25in}
    \caption{
    \textbf{Samples from the auto-labeling pipeline.}
    Each frame is annotated with three training labels produced by our auto-labeling pipeline:
    (1) the texting command expressed as a short verb phrase (e.g., ``go straight'', ``slow down'', ``speed up''),
    (2) the drafting, visualized as a rendered path from the ground-truth future trajectory, and
    (3) the arrowing input, represented by instantaneous speed.
    The rightmost panel shows the reasoning traces used to supervise drafting and arrowing prediction.
    }
    \label{fig:dataset anno}
    \vspace{-15pt}
\end{figure}

As illustrated in Figure~\ref{fig:dataset anno}, 
we provide examples of the multi-modal instructions generated by our labeling pipeline. It follows a two-stage strategy to determine which frames should be annotated. First, a pre-trained VLM (InternVL3-8B~\cite{zhu2025internvl3}) scores video frames based on visual complexity, such as pedestrian interactions, obstacles, or terrain variations, producing an “interestingness” prior. Next, motion statistics (e.g., acceleration and turning rate) are computed within sliding windows and fused with the VLM scores to obtain weighted motion saliency. Frames are then ranked using this combined score, ensuring that those with rich dynamics or meaningful interactions are prioritized. Finally, multimodal human instructions are synthesized from ground-truth trajectories, and Qwen2.5VL-72B~\cite{bai2025qwen2} is used to generate corresponding textual prompts.

After selecting informative frames, we generate three complementary types of annotations to provide rich supervisory signals for navigation learning.
Building such supervision at scale is non-trivial: each training sample must align (i) noisy real-world robot trajectories, (ii) calibrated camera geometry for accurate projection, and (iii) intent descriptions that are consistent with both the scene context and future motion. Importantly, these annotations are designed to mirror the human interfaces in shared autonomy: they let users intervene at different abstraction levels (sketching a route, nudging velocity, or stating intent) without requiring continuous low-level teleoperation. To balance clarity in the main paper and reproducibility, we summarize each modality below and defer implementation details (e.g., projection, sampling, prompting) to the Appendix~\ref{appendix:multimodal-anno}.

\noindent
\cmdtag{\texttt{Drafting}}\; A user can quickly \emph{draw a rough path} on the live observation to indicate where the robot should go (e.g., ``go around the group'' or ``take the right side''), which is often easier than issuing continuous joystick commands. We therefore represent this interface as a trajectory overlay with sampled pixel points, providing explicit spatial guidance for the policy.

\noindent
\cmdtag{\texttt{Arrowing}}\; When only a brief correction is needed (e.g., slow down, gently turn left), a low-bandwidth arrowing signal is a natural and lightweight intervention. We encode this interface with compact speed and heading supervision (and its visualization), enabling the model to react to short corrective nudges while still handling low-level stabilization.

\noindent
\cmdtag{\texttt{Texting}}\; Natural language is convenient for expressing high-level intent and constraints (e.g., ``yield to pedestrians'' or ``stay close to the curb''), especially under limited attention or communication latency. We use a vision-language model (Qwen2.5-VL-72B~\cite{bai2025qwen2}) to generate a command-style instruction plus a longer description, training the policy to align execution with human intent and scene interactions.

\subsection{Model Training}
\label{sec:Training strategy}
We employ a two-stage training strategy to efficiently learn the share-autonomy model.
\vspace{-5pt}

\paragraph{Instruction-conditioned VLM Adaptation.}
In the first stage, we adapt the InternVL3-2B~\cite{zhu2025internvl3} to incorporate human instruction conditioning. We freeze the vision encoder and the original vision-to-language projection MLP, and only train the newly introduced {\CmdEncodershort} modules.
Additionally, we employ LoRA~\cite{hu2022lora} for efficient adaptation of the language model. The model is trained using a language modeling loss on the generated trajectory captions, enabling the VLM to understand and encode semantic-spatial instruction signals through natural language grounding.
\vspace{-5pt}

\paragraph{End-to-End Diffusion Policy Learning.}
In the second stage, we train the diffusion-based policy network while keeping the multi-modal encoder frozen.
The diffusion decoder and auxiliary encoders (goal, camera, trajectory anchor encoders) are trained from scratch. The training loss combines mode classification loss $\mathcal{L}_{cls}$ and trajectory regression loss $\mathcal{L}_{reg}$: $\mathcal{L} = \mathcal{L}_{cls} + \mathcal{L}_{reg}$, where $\mathcal{L}_{cls}$ selects the mode closest to GT via cross-entropy, and $\mathcal{L}_{reg}$ minimizes L2 distance between the predicted and GT trajectories.
\vspace{-5pt}

\section{Experiments}
\begin{figure*}[t!]
    \centering
    \includegraphics[width=1\linewidth]{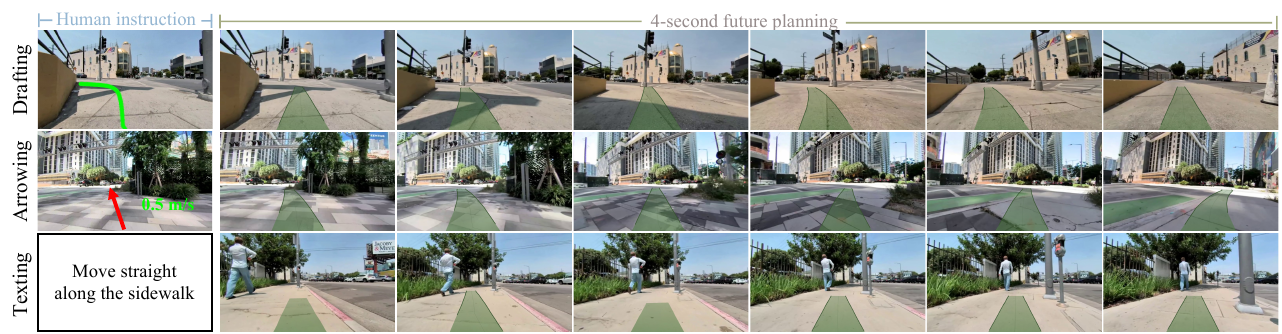}
    \caption{
\textbf{Visualization of offline inference in \textit{MM-CoS}.} We illustrate three types of human instructions. The green polygon denotes the future trajectories predicted by {\ModelName}. 
    }
    \vspace{-15pt}
    \label{fig:visurbanwalks}
\end{figure*}

We evaluate {\ModelName} from three aspects: (i) its ability to follow human instructions (Section~\ref{exp:instruction following}), (ii) the efficiency of shared control (Section~\ref{exp:share control efficiency}) and (iii) the ablation study to evaluate the effectiveness of components in the framework and the dataset (Section~\ref{exp: abl}).
% \wayne{also validate others?}
Finally, we conduct a pilot study that deploys the system in real-world sidewalk environments (Section~\ref{exp:realworld}). Additional qualitative results are provided in Appendix~\ref{apd:morevis}.

\subsection{Validation on Instruction Following}
\label{exp:instruction following}

We first evaluate our approach on instruction-following in an open-loop manner, where predicted trajectories are compared against ground-truth future trajectories from the \textit{MM-CoS} test set. For evaluation, we adopt the standard open-loop metrics proposed in prior works~\cite{varadarajan2022multipath++, ettinger2021large}. During testing, we provide our model with different input modalities, including texting, drafting, arrowing, and point. Motivated by the limited exploration of human instruction in urban navigation, we also adopt point-goal~\cite{liu2025citywalker, hirose2025learning} and image-goal navigation models~\cite{shah2022gnm, shah2023vint, sridhar2024nomad} as baselines for comparisons, since they provide more precise guidance compared to instruction-based approaches. 
\begin{table}[h!]
\centering
\caption{{\textbf{Open-loop evaluation on \textbf{MM-CoS}.} * denotes models re-trained on our dataset.} 
}
\vspace{-0.5em}
\label{tab:openloop}
\resizebox{\linewidth}{!}{%
\begin{tabular}{l ccccc}
\toprule
 & minADE$_{1s}$~$\downarrow$ & minFDE$_{1s}$~$\downarrow$  & mAP~$\uparrow$ & L2$_{1s}$~$\downarrow$ &  L2$_{2s}$~$\downarrow$ \\
\midrule
GNM$^\ddagger$~\cite{shah2022gnm} &0.594 &0.988 &- &0.988 &-\\
ViNT$^\ddagger$~\cite{shah2023vint} &0.638 &1.056 &- &1.056 &-\\
NoMaD$^\ddagger$~\cite{sridhar2024nomad} &0.523 &0.858 &0.216 &1.072 &2.182\\
\midrule
MBRA~\cite{hirose2025learning} &0.617 &1.019 &- &1.019 &2.034\\
CityWalker~\cite{liu2024citywalker} &0.648 &1.125 &-&1.125 &-\\

\midrule
ViNT*~\cite{shah2023vint} &0.247 &0.450 &- &0.425 &0.925\\
CityWalker*~\cite{liu2024citywalker} &0.180 &0.353 &-&0.353 &0.786\\

\midrule
\ours \ModelName~\includegraphics[height=0.1in]{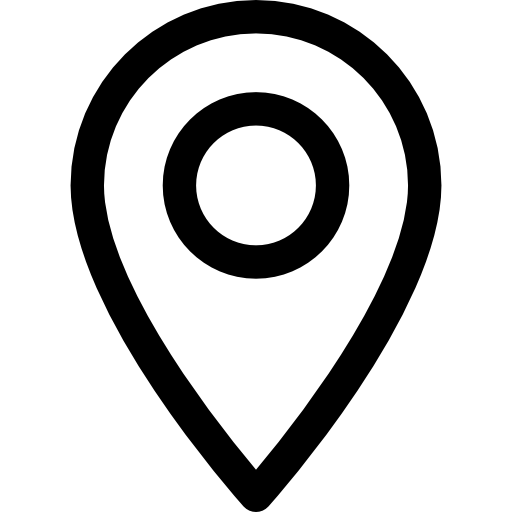} & {{0.125}} & {{0.218}} & {0.699} & {0.266} & {0.670} \\ 
\ours \ModelName~\includegraphics[height=0.1in]{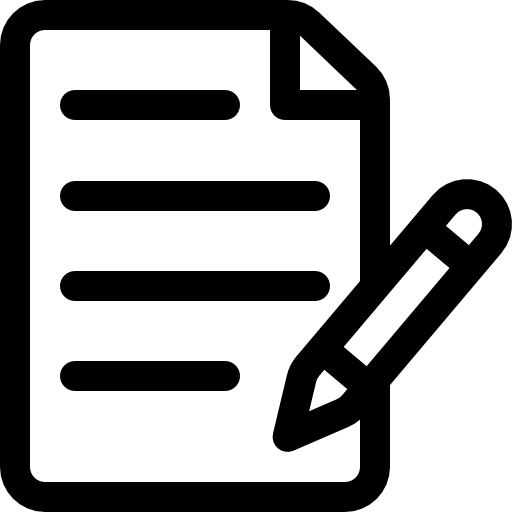} & 0.125 & 0.238 & 0.683 & 0.259 & 0.673 \\ 
\ours \ModelName~\includegraphics[height=0.1in]{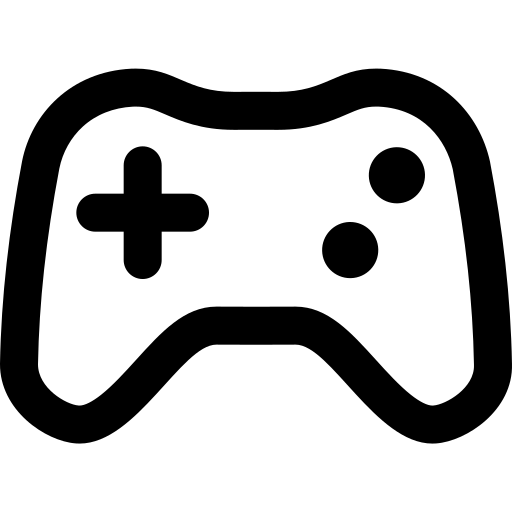} & {0.108} & 0.220 & 0.750 & {0.150} & {0.473} \\ 
\ours \ModelName~\includegraphics[height=0.1in]{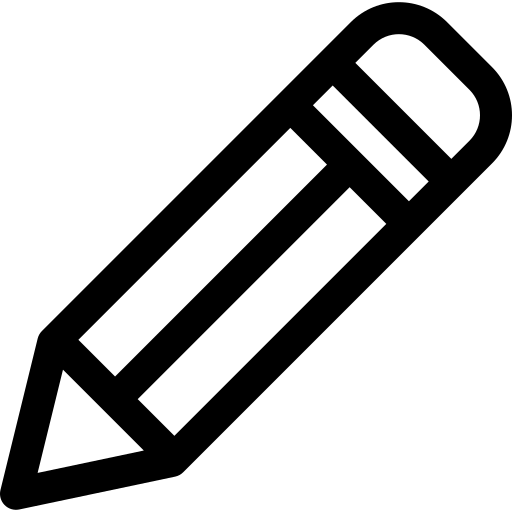} & 0.122 & 0.218 & {0.844} & {0.244} & {0.557} \\ 
\bottomrule

\end{tabular}
}
\vspace{-10pt}
\end{table}

Table~\ref{tab:openloop} shows that {\ModelName} consistently outperforms all baselines across instruction modalities. In particular, the arrowing-guided variant~\includegraphics[height=0.1in]{figs/icons/game.png} achieves the lowest L2 error (0.150 at 1s and 0.473 at 2s), outperforming CityWalker* (0.353 at 1s and 0.786 at 2s) by 39.8\% at 2s. The drafting-guided version~\includegraphics[height=0.1in]{figs/icons/pencil.png} yields the best mAP (0.844) while maintaining strong L2 performance (0.557 at 2s), whereas the text-guided version~\includegraphics[height=0.1in]{figs/icons/notes.png} is slightly worse in both mAP and L2. Overall, geometric instructions (drafting and arrowing) provide stronger spatial guidance for precise trajectory generation than purely linguistic instructions.
Some baselines omit mAP due to deterministic single-trajectory outputs, and omit L2$_{2s}$ due to shorter prediction horizons.

We provide qualitative results in Figure~\ref{fig:visurbanwalks}, illustrating how {\ModelName} responds to three distinct categories of human instructions. Under drafting or arrowing input as instruction, the model produces the geometrically aligned predictions in different scenarios, closely matching the user intention. With texting instructions, {\ModelName} converts high-level semantic intent into a coherent motion plan that respects both the instruction behavior and the surrounding scene, \textit{i.e.}, following the sidewalk direction. These examples demonstrate the model’s ability to interpret diverse human input and produce safe, intent-aligned trajectories for urban navigation.

\subsection{Shared Control Efficiency Analysis}
\label{exp:share control efficiency}
We aim to evaluate the efficiency of different shared-control methods. Specifically, we examine how robust each agent is to noisy waypoints and quantify the amount of time an operator needs to spend on takeover.
\vspace{-5pt}
\paragraph{Pseudo-simulation testing for shared control.}
Closed-loop evaluation is feasible in simulation or the real world but hard to scale because human takeovers add randomness, so we use a pseudo-simulation shared-control testing pipeline.
The pipeline includes (i) a navigation model that takes sequential video frames and noisy target points (random angle in $[-90^{\circ}, 90^{\circ}]$ and distance in $[0,10]\,\mathrm{m}$) to predict trajectories and (ii) a judgment module that decides per frame whether takeover is needed and records intervention frames (Appendix~\ref{apd:judgement module}).

\paragraph{Results.}
\begin{figure}[h!]
    \centering
    \includegraphics[width=1\linewidth]{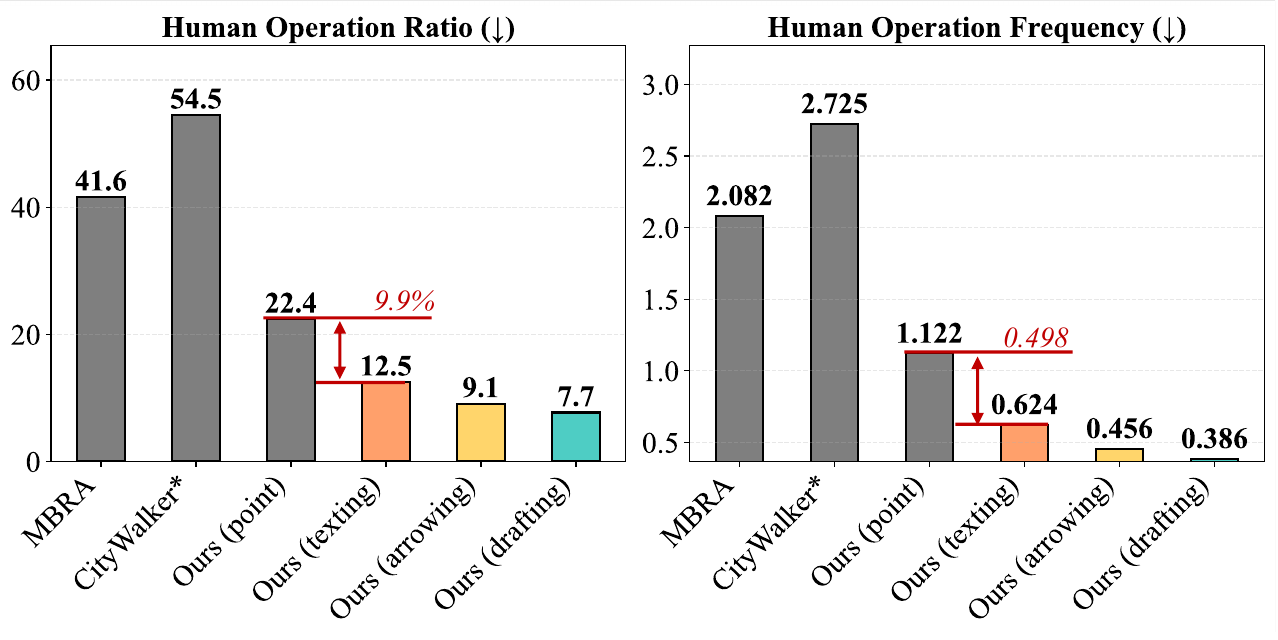}
    \caption{\textbf{Human Cost Evaluation in Pseudo-simulation.}
We compare the human intervention cost of our model against prior methods~\cite{mirowski2016learning, liu2024citywalker}, as well as across different modes of instruction guidance within our own framework.}
    \vspace{-5pt}
    \label{fig:sharecontrolbenchresults}
\end{figure}
We report two key metrics: Human Operation Ratio (percentage of time under human control) and Human Operation Frequency (number of frames under human control per total duration).
We report two key metrics: Human Operation Ratio (percentage of time under human control) and Human Operation Frequency (frames under human control per total duration). Results are shown in Figure~\ref{fig:sharecontrolbenchresults}, assuming each human takeover lasts 2 seconds. Two observations follow. First, our method achieves a human operation frequency of 0.96 and a 19.2\% lower Human Operation Ratio compared to the baseline~\cite{peng2023learning}, indicating improved robustness in point navigation. Second, when humans provide only high-level instructions, they require fewer control interventions, yielding a 9.9\% reduction in human operation time and a human operation frequency of 0.498 (44\% reduction). During testing, we use a 5Hz control rate, so the maximum human operation frequency is 5. More results for other takeover durations are in the Appendix~\ref{apd:share control}.

\begin{table}[h!]
\centering
\caption{\textbf{Evaluation on instruction following.}}
\label{tab:understanding}
\resizebox{\linewidth}{!}{%
\begin{tabular}{lcccccc}
\toprule
\textbf{Model} & 
\makecell{\textbf{Finetune}} & 
\makecell{\textbf{Visual} \\ \textbf{Prompt}} & 
\makecell{\textbf{\CmdEncodershort}} & 
\makecell{\textbf{ROUGE-L} $\uparrow$} & 
\makecell{\textbf{Intent} \\ \textbf{Score}  $\uparrow$}\\
\midrule
InternVL3-2B & $\times$ & \checkmark & $\times$ & 0.167 & 2.019 \\
InternVL3-8B & $\times$ & \checkmark & $\times$ & 0.184 & 2.818  \\
\midrule
InternVL3-2B & \checkmark & \checkmark & $\times$  & 0.532 & 4.885 \\
\ours {\ModelName} & \checkmark & $\times$ & \checkmark & 0.534 & 4.842 \\
\ours {\ModelName} & \checkmark & \checkmark & \checkmark & \textbf{0.581} & \textbf{5.446} \\
\bottomrule
\end{tabular}%
}
\end{table}

\subsection{Ablation Study}
\label{exp: abl}

To valid the dataset and the {\CmdEncodershort} we designed. 
To validate the effectiveness of the collected dataset and the proposed Instruction Encoder, we conduct ablation studies on both instruction understanding and planning capabilities.

\paragraph{Instruction following.}
We evaluate the model’s ability to reason correctly based on human instructions. Specifically, we report ROUGE-L~\cite{lin2004rouge} and Qwen Score. The Qwen Score is computed using QwenVL2.5-72B~\cite{bai2025qwen2}, a vision-language evaluation model. These metrics measure the quality of the model’s reasoning outputs compared to human-provided references.
The results are presented in Table~\ref{tab:understanding}. After fine-tuning on our synthesis label dataset, the InternVL3-2B model demonstrates significantly improved instruction following compared to its non-fine-tuned counterpart, and even surpasses the larger InternVL3-8B model in both accuracy and fluency. Using only the proposed projector module, the model achieves comparable performance to the original VLM with visual prompting. When combining both visual prompts and the projector, the model attains the best overall performance in instruction following.

% \vspace{-5pt}
\paragraph{End-to-end planning.}
To evaluate the effects of semantic and spatial awareness in our {\CmdEncodershort}, we conduct trajectory planning experiments under human-provided drafting and arrowing inputs. The horizontal axis denotes the time lag between each instruction and the current frame, with larger values indicating older instructions.  We compare four models: the baseline with only the DiT action decoder, where observations and the instruction image are encoded with Dinov3~\cite{simeoni2025dinov3} using visual prompts; {\ModelName} without geometry, which uses only visual prompts; {\ModelName} without without semantic, which directly embeds the drafting and arrowing inputs without any visual prompts; and {\ModelName} with both semantic and geometry.  
\begin{figure}[t!]
    \centering
    \includegraphics[width=1\linewidth]{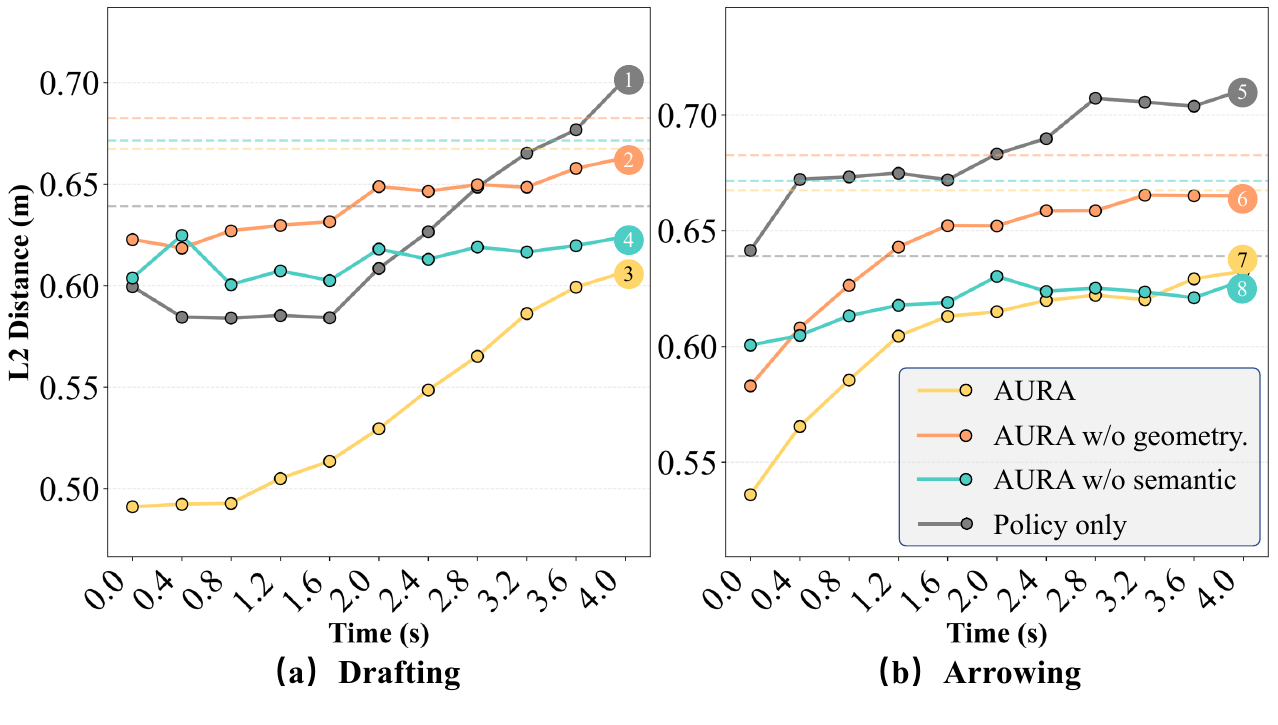}
    % \vspace{-0.15in}
    \caption{\textbf{Effectiveness of components on end-to-end planning.}}
    % \vspace{-5pt}
    \label{fig:abl}
\end{figure}
\begin{table}[t]
\centering
\caption{\textbf{Real-world closed-loop navigation results.}
We report Human Operation Ratio (HO), Normalized Intervention Rate (NIR, emergency interventions per 100 meters), Off-track Distance Ratio (ODR), and Time Success Rate (TSR, ratio of valid autonomous time to total autonomous time). Lower is better for HO/NIR/ODR and higher is better for TSR.}
\label{tab:realworld-nav}
{\scriptsize
\setlength{\tabcolsep}{8pt} 
\renewcommand{\arraystretch}{0.95} 
\resizebox{\linewidth}{!}{
\begin{tabular}{l cccc}
\toprule
Method & HO (\%)~$\downarrow$ & NIR~$\downarrow$ & ODR~$\downarrow$ & TSR~$\uparrow$ \\
\midrule
NoMaD & 9.74 & 43.2 & 11.3 & 89.0 \\
CityWalker & 14.56 & 48.29 & 20.0 & 80.3 \\
Gemini & 16.9 & 255.7 & 32.0 & 63.2 \\
\midrule
\ModelName & \textbf{1.73} & \textbf{16.99} & \textbf{10.5} & \textbf{89.3} \\
\bottomrule
\end{tabular}
}
}
\vspace{-5pt}
\end{table}

% \vspace{-5pt}
Figure~\ref{fig:abl} visualizes the L2 distance at 2s for these models. Dashed lines indicate the corresponding results without any target input. Line \circnumwhite{colgray}{1}
and \circnumwhite{colgray}{3} demonstrates the importance of semantic understanding: lacking a shared semantic representation, the model fails to interpret human instruction and performs worst when predicting instructions 4s ahead, even worse than the model without any target input. Comparing lines \circnumwhite{colorange}{2} and \circnumwhite{colgreen}{4}, and lines \circnumwhite{colorange}{6} and \circnumwhite{colgreen}{8}, we observe that geometry encoding provides similar short-term performance as visual prompts, since both convey goal direction information. However, over longer horizons, geometry encoding establishes a stable spatial structure and goal memory, whereas the explicit information in visual prompts disappears as the field of view changes, leading to performance degradation. Combining both semantic and geometric guidance achieves the best overall performance (lines \circnumwhite{colyellow}{3} and \circnumwhite{colyellow}{7}), as visual prompts ensure accurate short-term tracking while geometry encoding maintains long-term spatial consistency and goal-directed memory.

\subsection{Real-World Pilot Study}
\label{exp:realworld}
We conduct a pilot study by deploying the proposed shared-autonomy system on a wheeled robot in real-world sidewalk environments. We evaluate 8 scenarios covering 16 routes with a total length of about 2.8 km.

\begin{figure}[t!]
    \centering
    \includegraphics[width=1\linewidth]{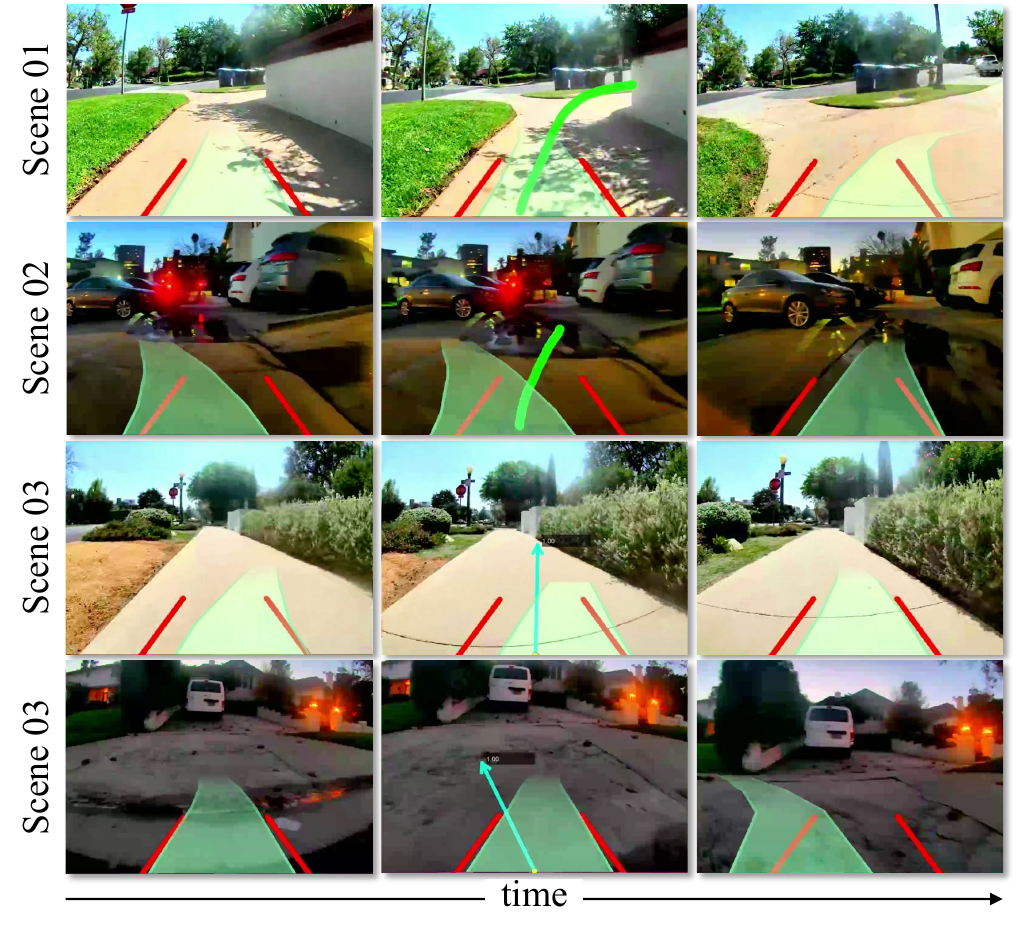}
    \vspace{-15pt}
    \caption{\textbf{Visualization of the teleoperation interface in real-world experiments.} 
For each row, the center frame corresponds to the moment of human intervention (teleoperation takeover), with the other frames showing the surrounding context. The green polygon denotes the predicted trajectories, and the red lines on both sides indicate the width of the robot.}
\vspace{-5pt}
    \label{fig:realworld}
\end{figure}

As shown in Table~\ref{tab:realworld-nav}, we compare against NoMaD, CityWalker, and Gemini, using four metrics tailored to shared autonomy: Human Operation Ratio (HO), Normalized Intervention Rate (NIR, emergency interventions per 100 meters), Off-track Distance Ratio (ODR), and Time Success Rate (TSR, ratio of valid autonomous time to total autonomous time). With safety-supervised human interventions, all routes can be completed; thus HO/NIR quantify human cost. {\ModelName} achieves the lowest human intervention cost across all metrics.
Figure~\ref{fig:realworld} presents qualitative examples from the real-world experiments. Additional visualizations are provided in the Appendix~\ref{apd:morevis}.

\section{Conclusion}
We present an assistive robot autonomy system with hierarchical takeover to reduce effort. It pairs a multimodal encoder that aligns with human instructions with a diffusion policy for planning. Extensive experiments have validated the improved instruction-following capability and the benefits of shared control. 

\paragraph{Acknowledgment} The project was supported by the NSF Grants CCF-2344955 and IIS-2339769. Honglin is supported by the Amazon Trainium Fellowship. The project and paper were finished during Yukai Ma's visit to UCLA.

\small
\bibliographystyle{ieeenat_fullname}
\bibliography{main}

\newpage
\appendix

\begin{figure*}[hbt!]
    \centering
    \includegraphics[width=\textwidth]{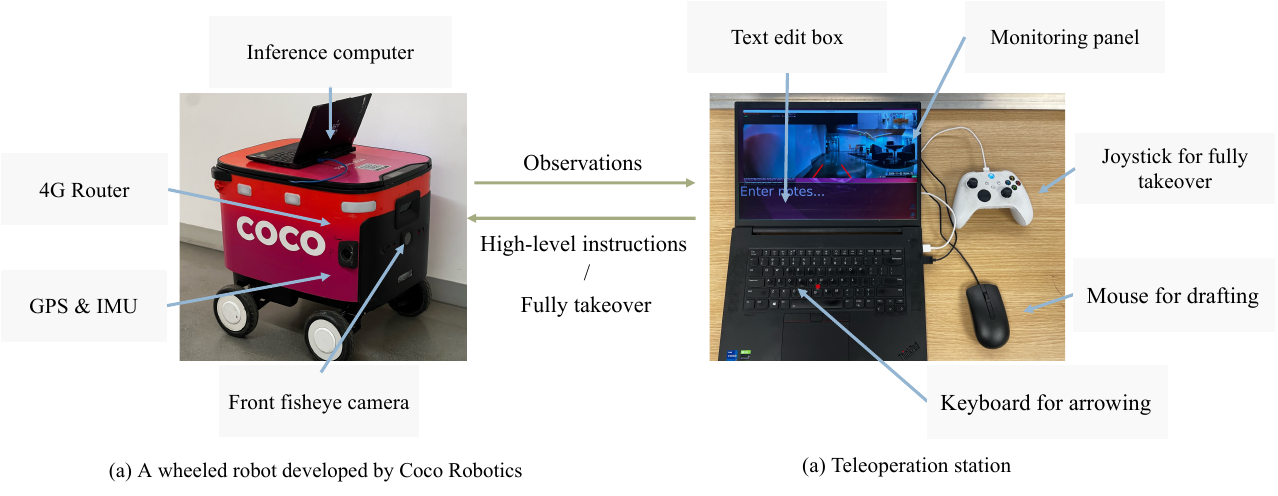}
    \caption{\textbf{Overview of the real-world experiment setup.} Left: the robot hardware platform; Right: the teleoperation interface.}
    \label{fig:coco}
\end{figure*}

\begin{center}
    \LARGE \textbf{Appendix}
\end{center}

In the appendix, we provide additional details of this work.
In Section~\ref{apd:real world}, we introduce the robot and teleoperation platform, along with more qualitative results.
In Section~\ref{sec:more detail about urbanwalks}, we provide further details about our \textit{MM-CoS} dataset, including data distribution and the automatic labeling pipeline.
In Section~\ref{apd:sharecontrol}, we present additional quantitative results on efficiency under \texttt{Share Control}.
In Section~\ref{sec:implementation_details}, we describe the implementation details.

{
\startcontents[sections]
\printcontents[sections]{l}{1}{\setcounter{tocdepth}{3}}
}
\section{Real-World Experiments}
\label{apd:real world}

\subsection{Evaluation Platform}
We evaluate the instruction-following performance of our model on a wheeled robot developed by \href{https://www.cocodelivery.com/}{Coco Robotics}. As illustrated in Figure~\ref{fig:coco}, the robot platform includes both the onboard robotic infrastructure and a teleoperation interface for monitoring and control. Notably, during testing, the inference computer is placed inside the robot’s storage compartment.
For the onboard robotic system, we primarily use the front fisheye camera for perception. The raw fisheye images are undistorted before being passed to the inference computer. The robot is connected to the inference computer through an Ethernet cable, and the inference computer outputs the desired linear and angular velocities generated by a PD controller, which are then sent to the robot executor. The robot performs localization using odometry based on IMU and GPS fusion. Communication with the teleoperation station is established through a 4G router.
The teleoperation station consists of a computer equipped with a joystick, mouse, and keyboard. The mouse and keyboard are used for issuing instructions, while the joystick is reserved for full manual control.

\subsection{More Visualization Results.}
\label{apd:morevis}
\begin{figure*}[t]
    \centering
    \includegraphics[width=0.95\linewidth]{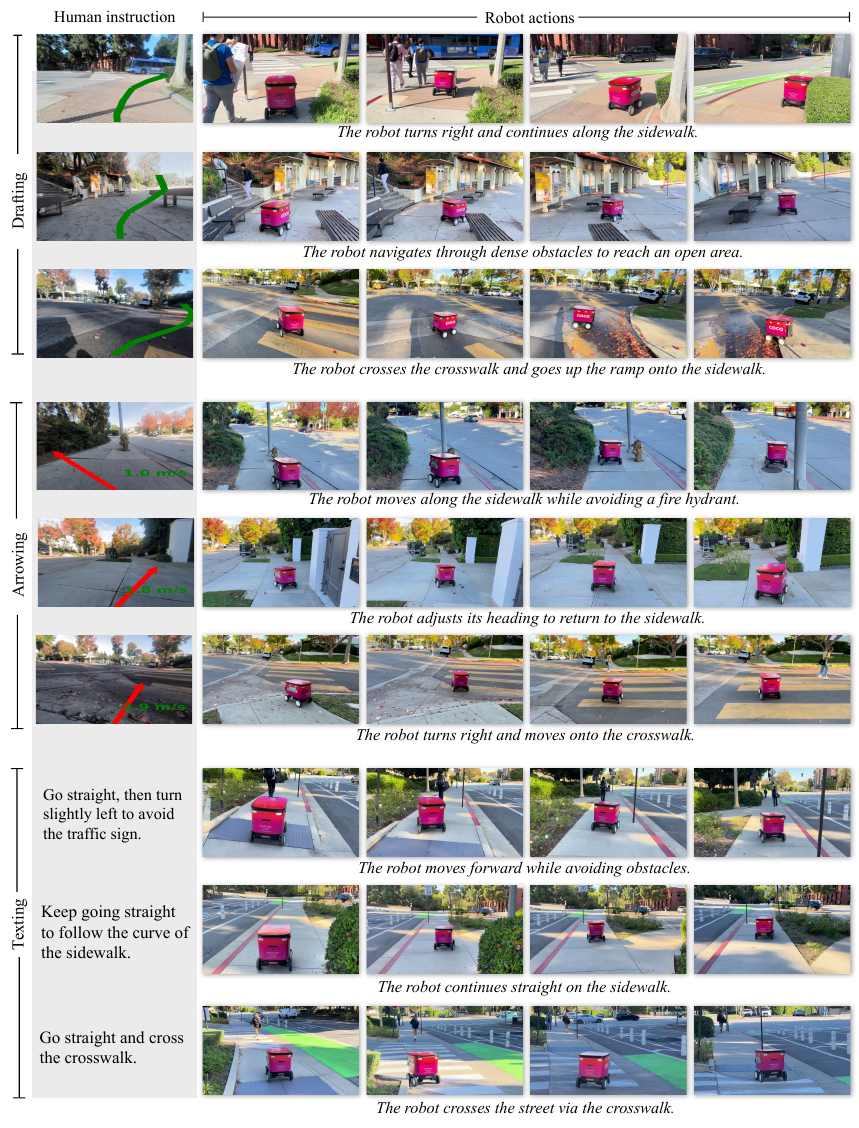
    }
    % \vspace{-0.15in}
    \caption{
    \textbf{Qualitative results in the real world.} The first three rows are guided by drafting, rows 4 to 6 by arrowing, and the last three rows by texting. The first column shows human instructions, while columns 2 to 5 show the actions executed by the robot.
    }
    % \vspace{-0.20in}
    \label{fig:real world demo}
\end{figure*}

We present additional real-world results in Figure~\ref{fig:real world demo}, showcasing three cases for each type of instruction. {\ModelName} demonstrates robust performance in real-world navigation tasks, including path selection, lane recovery, and obstacle avoidance. The captions under the images describe the robot's actions. The first three rows illustrate that the robot can accurately interpret geo-information under drafting guidance. Rows four to six show that the robot can follow arrowing instructions, effectively performing obstacle avoidance, lane recovery, and path selection. Finally, rows seven to nine demonstrate that our model can follow high-level textual instructions provided by a human.

\section{Details about \textit{MM-CoS}.}
\label{sec:more detail about urbanwalks}
\subsection{\textbf{\textit{MM-CoS}}}
\label{sec:urbanwalks}
\begin{figure*}[h!]
    \centering
    \includegraphics[width=1\linewidth]{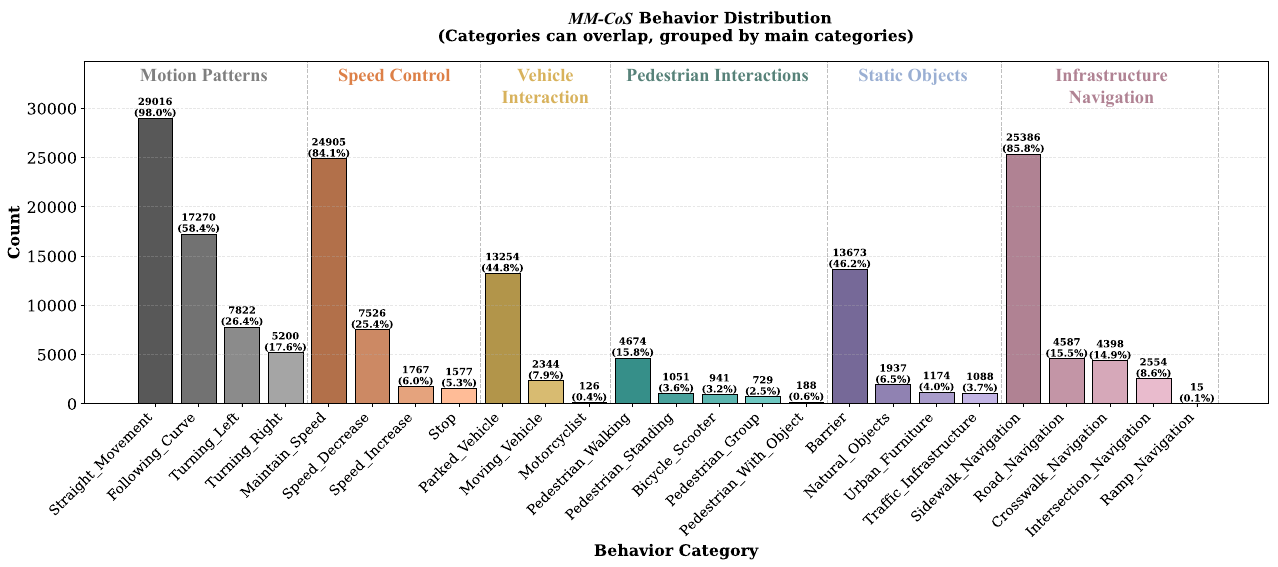}
    \caption{
    \textbf{Data distribution of our \textit{MM-CoS} dataset (50h)}. Behaviors are divided into six subclasses based on their outcomes and causes. Bars of the same color represent different specific actions within each subclass. The top of each bar indicates the number of occurrences of the behavior and its proportion of the total annotations.
    }
    \label{fig:data distribution}
\end{figure*}

\begin{figure*}[h!]
    \centering
    \includegraphics[width=1\linewidth]{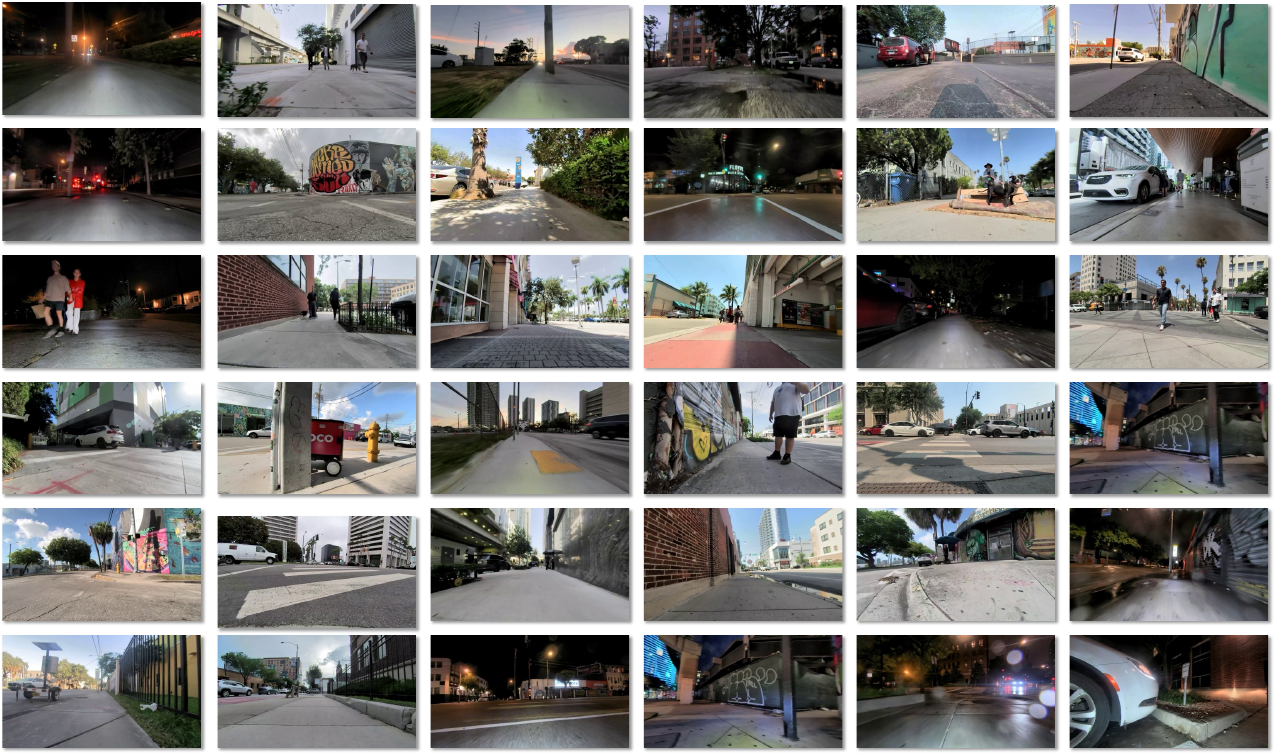}
    \caption{
    \textbf{A thumbnail montage showing a subset of the \textit{MM-CoS} videos collected from real world environments for shared autonomy.
    }
    }
    \label{fig:cocodata sample}
\end{figure*}
\begin{figure}[t!]
    \centering
    \includegraphics[width=1\linewidth]{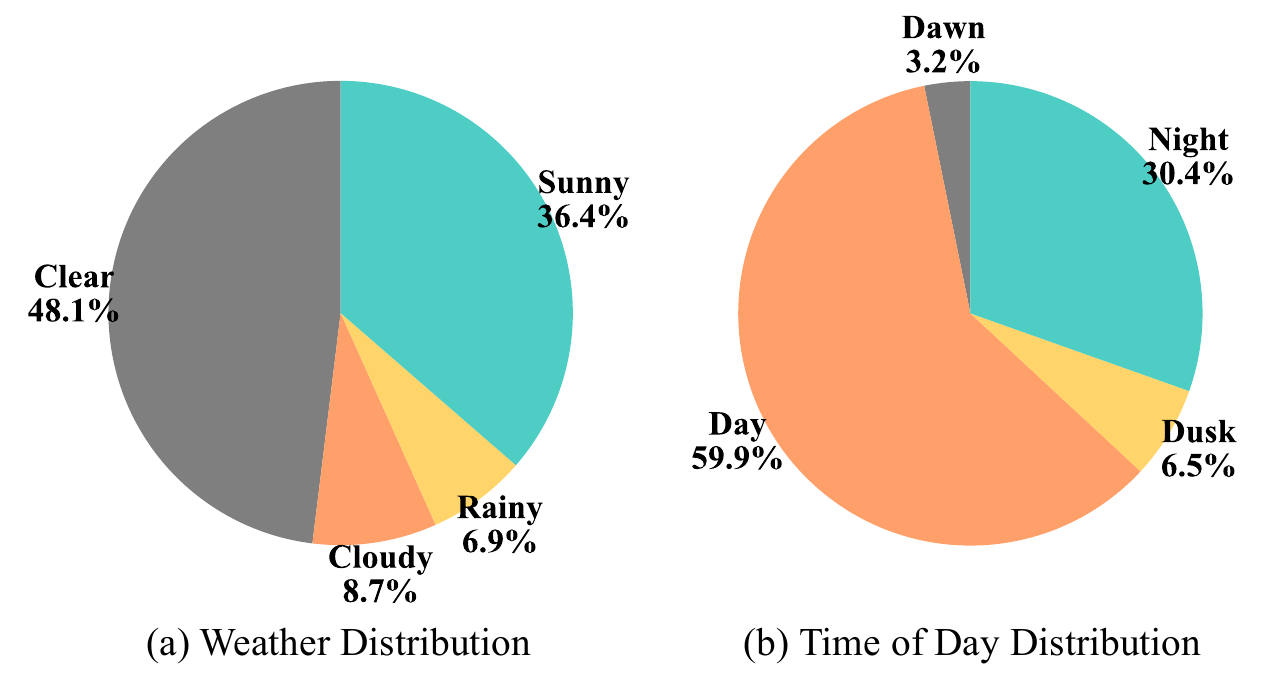}
    % \vspace{-0.25in}
    \caption{
     \textbf{Data distribution across weather conditions and time of day.} The dataset contains 3,040 scenes in total, with weather conditions primarily consisting of clear (48.1\%) and sunny (36.4\%) conditions, while cloudy (8.7\%) and rainy (6.9\%) conditions are less represented. Time-of-day distribution shows a strong bias toward daytime scenes (59.9\%), followed by nighttime scenes (30.4\%), with transitional periods of dawn (3.2\%) and dusk (6.5\%) being relatively scarce.
}
    \label{fig:dataweatherdistribution}
\end{figure}

Finally, we generate 29K annotations for our \textit{MM-CoS} dataset from 50 hours of teleoperation data. Figure~\ref{fig:data distribution} illustrates the data distribution. The frequencies were calculated based on keyword occurrences, and the data are categorized by action mode and interaction behavior. Each major category contains specific subcategories, which may overlap; for example, a single scene could include both ``Straight\_Movement'' and ``Turning\_Left'' at the same time. The percentage values shown on each bar indicate the proportion of the dataset corresponding to that behavior. The weather distribution and time of day are illustrated in Figure~\ref{fig:dataweatherdistribution}. The \textit{MM-CoS} dataset (50 hours) contains 3,040 videos, covering a wide range of weather and lighting conditions.
Figure~\ref{fig:cocodata sample} illustrates the diversity of the \textit{MM-CoS} dataset, covering variations in lighting, weather, and sidewalk scenes. Such data better reflect real-world scenarios and present significant challenges.

In addition, the \textit{MM-CoS} dataset is further augmented with 18K annotations from open-source datasets~\cite{shah2021rapid, karnan2022socially, akhtyamov2025egowalk}.
% \subsection{Auto labeling pipeline}

\subsection{Auto labeling pipeline}
\label{appendix:label pipeline}
\begin{figure}[t!]
    \centering
    \includegraphics[width=1\linewidth]{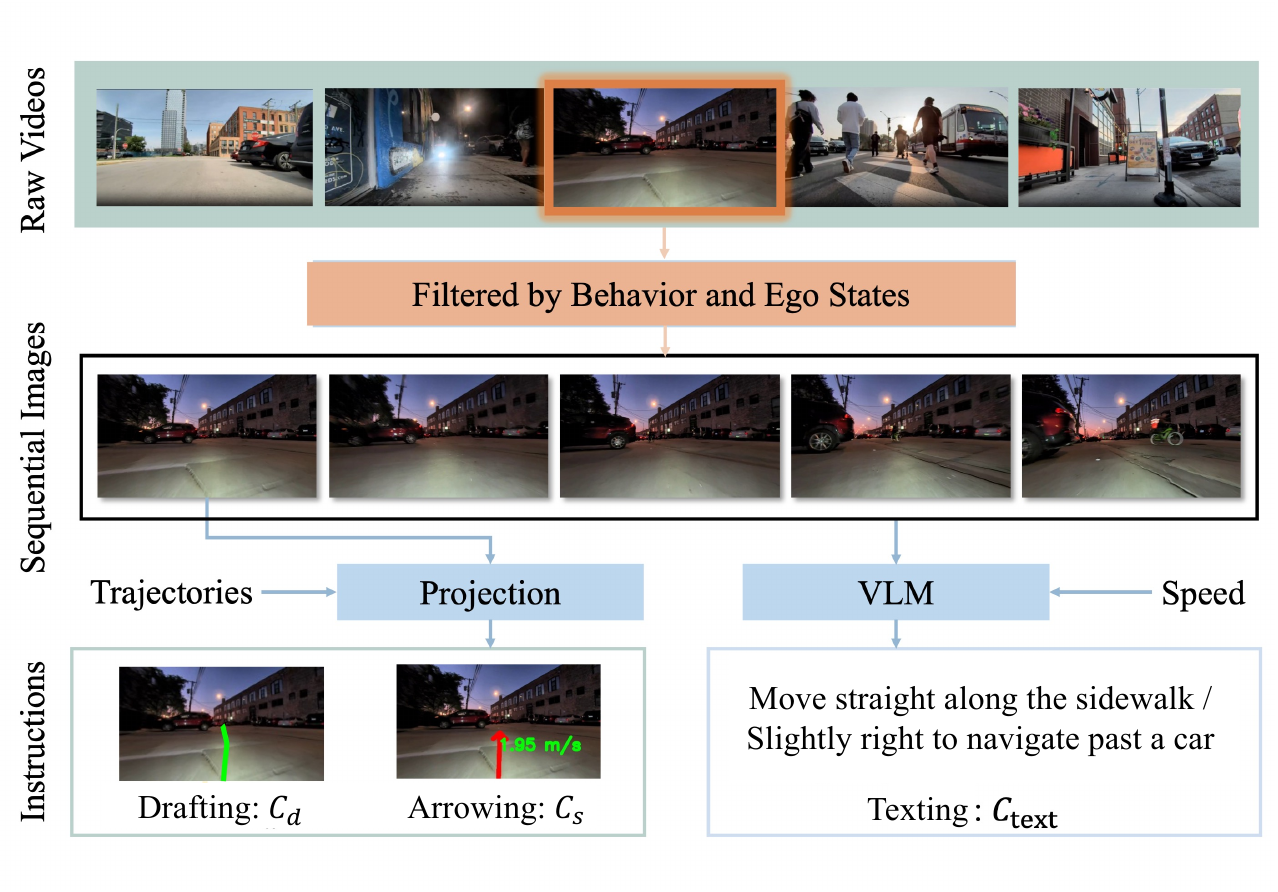}
    \caption{
    \textbf{Auto labeling pipeline.} 
We first select interesting clips based on behavioral cues and ego state information, focusing on frames that show rich interactions with objects or the environment. We then feed the front view images with visual prompts of the projected future trajectory and the speed of each frame into the VLM to generate textual instructions. Finally, we use the projected future trajectory on the front images to derive the drafting and arrowing instructions.
    }
    \label{apfig:dataset anno}
\end{figure}
As illustrated in Figure~\ref{apfig:dataset anno}, we employ a two-stage selection strategy to identify informative frames for annotation. This approach combines VLM-based assessment with trajectory-based motion analysis, corresponding to the “behavior and state filtering” step shown in the figure.

\begin{figure}[h!]
    \centering
    \includegraphics[width=1\linewidth]{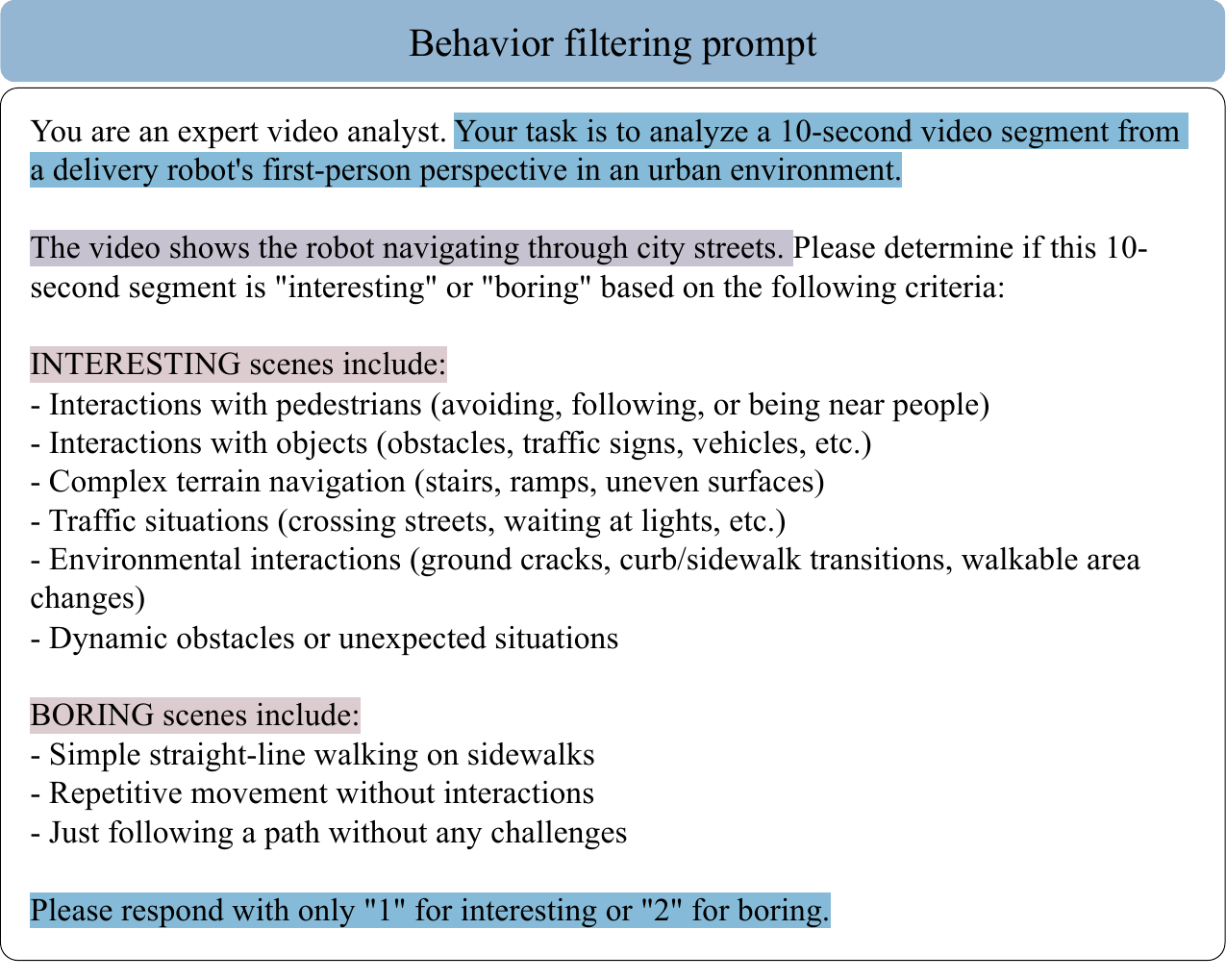}
    \caption{
\textbf{Prompt for behavior filter.} We use this prompt to let the VLM determine whether the robot in the clip is interacting with objects or the environment.
    }
    \label{fig:interesting prompt}
\end{figure}

First, we use a pre-trained VLM (InternVL3-8B~\cite{zhu2025internvl3}) to classify each video segment as either “interesting’’ or “boring’’ based on scene complexity, such as pedestrian interactions, obstacles, and terrain changes. This produces a temporal interestingness map $\mathbf{I} = [i_1, i_2, \ldots, i_{n}] \in {1, 2}^{n}$ for the $n$ segments of each clipped trajectory, where $i_j = 1$ denotes an interesting segment and $i_j = 2$ denotes a boring one. The prompts used for this classification are shown in Figure~\ref{fig:interesting prompt}.

Second, we perform motion analysis on the robot trajectory using a sliding window approach. For each window, we compute acceleration and turning scores as weighted combinations of motion statistics: \(S_{accel} = w_1 \alpha_{avg} + w_2 \alpha_{max} + w_3 \sigma_\alpha^2\) and \(S_{turn} = w_4 \theta_{avg} + w_5 \theta_{max} + w_6 \sigma_\theta^2\), where \(\alpha\) and \(\theta\) denote acceleration and turning angle respectively. To prioritize semantically meaningful scenes, we apply priority-based weighting: \(S' = w_p \cdot S\) where \(w_p = w_{\text{int}} > 1\) for interesting segments, \(w_p = w_{\text{bor}} < 1\) for boring segments, and \(w_p = 1\) otherwise. We then rank windows by their weighted scores and select the top-\(k\) frames with high acceleration or turning, ensuring temporal diversity by filtering adjacent candidates. 

\begin{figure}[h!]
    \centering
    \includegraphics[width=1\linewidth]{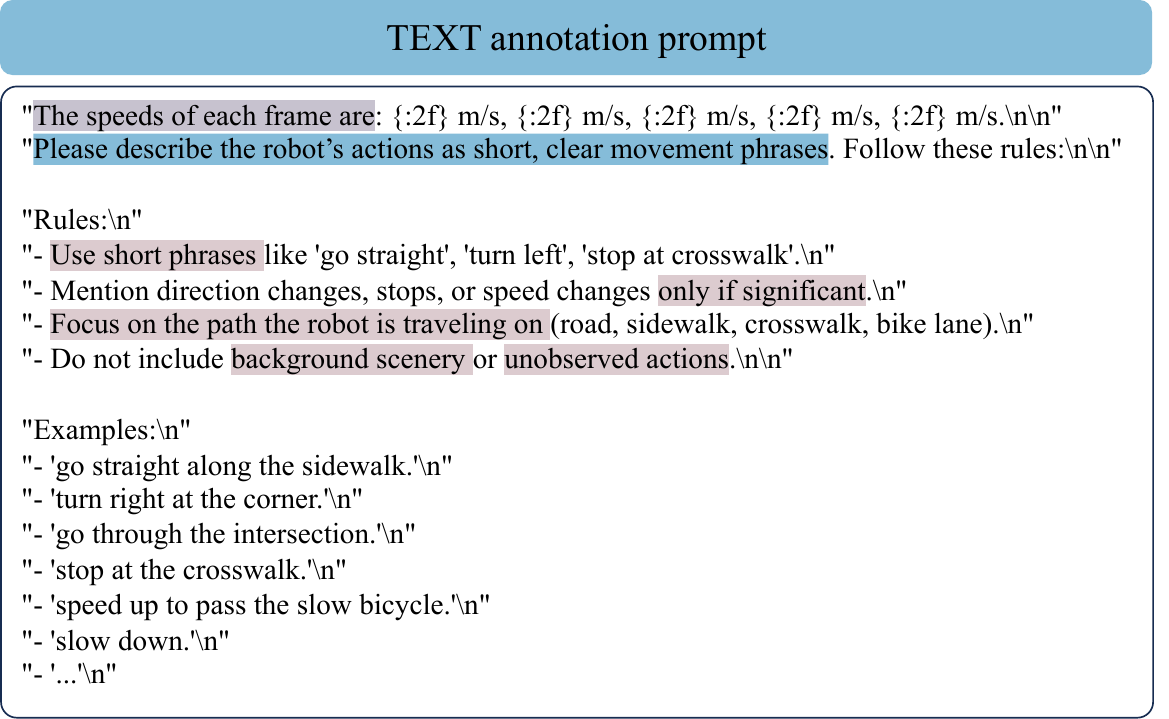}
    \caption{
 \textbf{Prompt for generating texting instructions.} We expect the VLM to produce short, simple instructions that guide the robot at a high level.
    }
    \label{fig:text anno prompt}
\end{figure}

\begin{figure}[h!]
    \centering
    \includegraphics[width=1\linewidth]{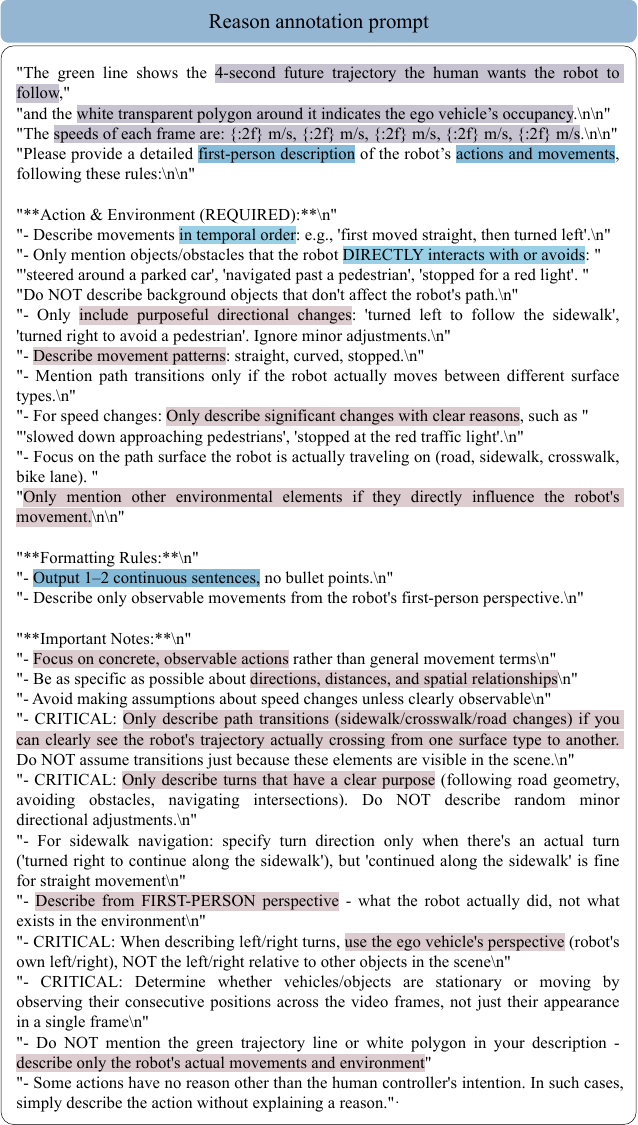}
    \caption{
 \textbf{Prompt for generating reasoning.} We expect the VLM to generate a detailed description of the robot’s behavior in the video, and to provide supervision signals (drafting and arrowing) for interpreting human instructions.
    }
    \label{fig:reason anno prompt}
\end{figure}

Once keyframes are identified (Figure~\ref{apfig:dataset anno}), we construct the VLM inputs by overlaying the robot’s future trajectory and frame-wise speed cues on the front-view images. With the prompts in Figures~\ref{fig:text anno prompt} and~\ref{fig:reason anno prompt}, the VLM generates (i) a short command-style instruction and (ii) a detailed description of the underlying reasons. In parallel, we derive geometric/control supervision (drafting and arrowing) from the same trajectory and speed signals; formal definitions are provided in Section~\ref{appendix:multimodal-anno}. Notably, for all appendix prompts, the purple text explains the input signals and the pink text specifies rules to follow so that the output satisfies the requirements shown in blue.

\subsection{Multi-Modal Annotation Generation}
\label{appendix:multimodal-anno}

For each selected frame, we generate three complementary types of annotations to provide rich supervisory signals for navigation learning.

\noindent
\cmdtag{\texttt{Drafting.}}
We project the robot's planned 4-second future trajectory from 3D world coordinates onto the 2D image plane using a pinhole camera model with intrinsic parameters $[f_x, f_y, c_x, c_y]$ and the corresponding 6-DoF camera poses. The projected waypoints are rendered as green trajectory lines overlaid on the RGB image $\boldsymbol{I}_d$, showing the intended path relative to the observed scene. Along this line, we uniformly sample $n$ pixel points $\boldsymbol{p}_d$ to obtain the visual instruction $C_d = (\boldsymbol{I}_d, \boldsymbol{p}_d)$.

\noindent
\cmdtag{\texttt{Arrowing.}}
We compute the average speed and directional control signals from the trajectory waypoints. Specifically, given the first $N$ waypoints $\{\mathbf{p}_i\}_{i=1}^N$, the average velocity vector is calculated as $\mathbf{v}_{\text{avg}} = \frac{1}{N-1} \sum_{i=1}^{N-1} (\mathbf{p}_{i+1} - \mathbf{p}_i) \cdot \text{fps}$, and the corresponding heading angle as $\theta = \arctan2(v_y, v_x)$. The velocity vector $\mathbf{v}_{\text{avg}}$ is then projected onto the front-view camera image and visualized as a fixed-length arrow indicating direction, with the absolute speed value overlaid as text, producing the image $\boldsymbol{I}_s$. The resulting arrowing instruction is defined as $C_s = (\boldsymbol{I}_s, \mathbf{v}_{\text{avg}})$.

\noindent
\cmdtag{\texttt{Texting.}}
We employ a pre-trained vision-language model, Qwen2.5-VL~\cite{bai2025qwen2}, deployed with vLLM~\cite{kwon2023efficient} for efficient inference, to generate two levels of language supervision: (i) a short command-style instruction (verb phrase) describing the maneuver (e.g., ``go straight'', ``slow down'', ``speed up''), and (ii) a more detailed natural-language description used to supervise reasoning about the underlying scene and interactions. We sample future observation frames covering 4 seconds and overlay the corresponding 4-second trajectory visualizations on each frame. The VLM processes these temporally ordered frames together with frame-wise speed measurements to produce the instruction and the detailed description, covering directional changes, speed adjustments, obstacle avoidance, and interactions with the environment. These language annotations provide high-level semantic insights into navigation behaviors, complementing the geometric and control-level annotations.

\section{Details in Shared Control}
\label{apd:sharecontrol}
\subsection{The Judgment Module.}

\label{apd:judgement module}
% \paragraph{}
\begin{figure}[h!]
    \centering
    \includegraphics[width=1\linewidth]{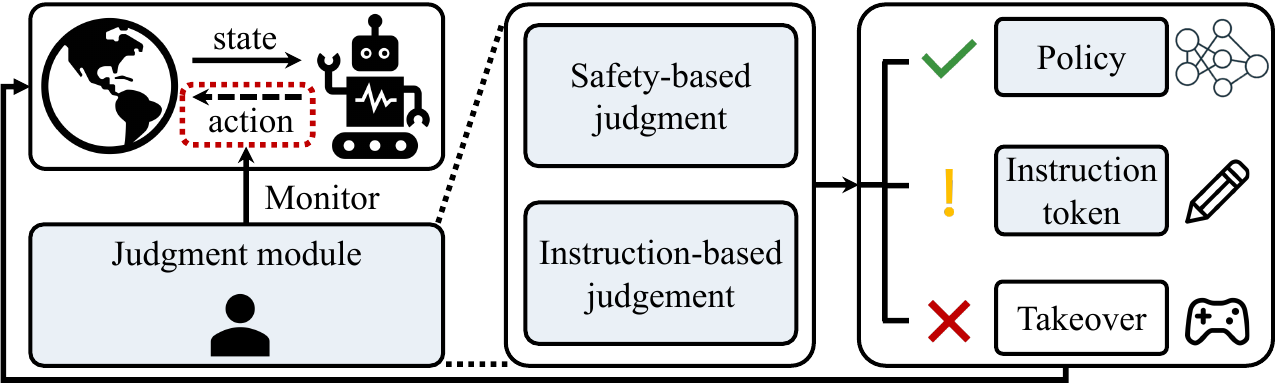}
    % \vspace{-0.15in}
    \caption{
    \textbf{Judgment module.} After the policy updates its state and receives the observation, it produces a candidate action through a forward pass of the model, which is temporarily cached before execution. We then apply a judgment module that evaluates the action using both safety-based criteria and instruction-based constraints. Based on this evaluation, the system decides whether to (i) execute the action, (ii) inject an instruction token to guide the policy toward human instruction, or (iii) trigger a full human takeover when the candidate action violates critical safety conditions.
    % \myk{mv to appendix}
    }
    % \vspace{-0.20in}
    \label{fig:sharecontrolbench}
\end{figure}
\begin{table*}[!t]
\centering
\caption{{\textbf{Pseudo simulation evaluation on share control}.}}
\vspace{-0.5em}
\label{tab:sharecontrol}
\resizebox{\linewidth}{!}{%
\begin{tabular}{l ccc|ccc|ccc|ccc}
\toprule
& HO\%~$\downarrow$ & TF~$\downarrow$ & OF~$\downarrow$
& HO$_{1s}$\%~$\downarrow$ & TF$_{1s}$~$\downarrow$ & OF$_{1s}$~$\downarrow$
& HO$_{2s}$\%~$\downarrow$ & TF$_{2s}$~$\downarrow$ & OF$_{2s}$~$\downarrow$
& HO$_{3s}$\%~$\downarrow$ & TF$_{3s}$~$\downarrow$ & OF$_{3s}$~$\downarrow$
 \\
\midrule
MBRA~\cite{hirose2025learning} & 15.1 & 0.757 & 0.757 & 30.5 & 0.305 & 1.523 & 41.6 & 0.211 & 2.082 & 48.1 & 0.164 & 2.407 \\
CityWalker*~\cite{liu2025citywalker} & 19.6 & 0.978 & 0.978 & 40.3 & 0.404 & 2.016 & 54.5 & 0.276 & 2.725 & 60.7 & 0.207 & 3.037 \\
\ours \ModelName~\includegraphics[height=0.1in]{figs/icons/maps-and-flags.png} & 6.0 & 0.298 & 0.298  & 14.9 & \textbf{0.151} & 0.744 & 22.4 & \textbf{0.116} & 1.122 & 28.0 & \textbf{0.098} & 1.400 \\
\midrule
\ours \ModelName~\includegraphics[height=0.1in]{figs/icons/notes.png} & 7.2 & 0.359 & 0.359 & 9.0 & 0.195 & 0.450 & 12.5 & 0.170 & 0.624 & 15.0 & 0.152 & 0.751 \\
\ours \ModelName~\includegraphics[height=0.1in]{figs/icons/game.png} & 5.0 & 0.248 & 0.248 & 6.4 & 0.169 & 0.320 & 9.1 & 0.152 & 0.456 & 11.2 & 0.144 & 0.562 \\
\ours \ModelName~\includegraphics[height=0.1in]{figs/icons/pencil.png} & \textbf{4.5} & \textbf{0.226} & \textbf{0.226} & \textbf{5.7} & {0.163} & \textbf{0.284} & \textbf{7.7} & {0.150} & \textbf{0.386} & \textbf{9.7} & 0.142 & \textbf{0.483} \\
\bottomrule
\end{tabular}%
}
\end{table*}

As shown in Figure~\ref{fig:sharecontrolbench}, we employ both rule-based and VLM-based criteria to determine whether a human takeover is required. There are two primary situations that trigger a takeover.
The first is collision risk. Specifically, we project the predicted trajectory together with the ego vehicle’s occupancy polygon onto the front camera view to obtain a mask $M_p$. Using OpenSeed~\cite{zhang2023simple}, we segment all pixels corresponding to impassable regions (such as obstacles or walls) as mask $M_f$. If the overlap between $M_p$ and $M_f$ exceeds 10\% of $M_p$, the system flags the frame as requiring a takeover.
The second is instruction compliance. To assess whether the predicted trajectory aligns with the human’s original instructions, we input two rendered images, one with the predicted trajectory and one with the ground-truth trajectory, into QwenVL2.5-72B~\cite{bai2025qwen2}. The model evaluates whether the prediction follows the intended goal, for example when the agent deviates from the correct path or remains stationary while the human intends to move.

\begin{figure}[h!]
    \centering
    \includegraphics[width=1\linewidth]{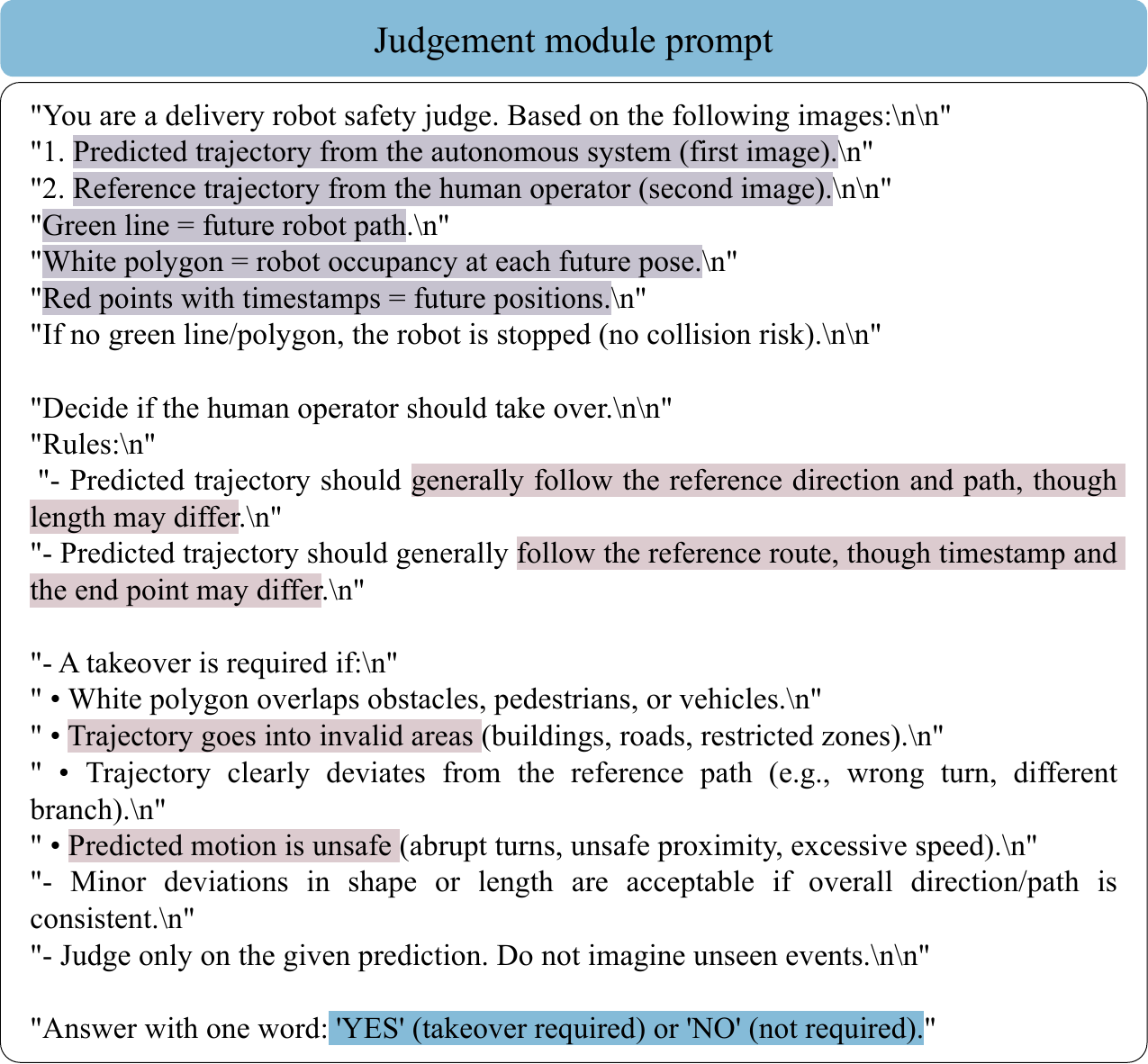}
    % \vspace{-0.15in}
    \caption{
\textbf{Prompt for the VLM judge.} The judge module is required to output only yes or no. It checks whether the predicted trajectory matches the human-intended action. If the robot's intent diverges from the human intention, human intervention is needed. For example, the robot may take an incorrect path or make unnecessary stops.
    % \myk{mv to appendix}
    }
    % \vspace{-0.20in}
    \label{fig:judge prompt}
\end{figure}

\subsection{More Results}
\label{apd:share control}

These experiments evaluate how much human intervention time can be reduced during teleoperation. To ensure reproducibility, we conduct all evaluations in a pseudo-simulation environment. We select 29 scenarios from the test set in which every clip is annotated as \textit{interesting}.

Additional results on Human Cost Evaluation in pseudo-simulation are shown in Table~\ref{tab:sharecontrol}. Here, HO denotes the human operation ratio, defined as the percentage of time under human control. TF represents the takeover frequency, calculated as the total number of takeover events divided by the total duration. OF indicates the operation frequency, defined as the number of operation actions per unit time. The first three columns without footnotes assume that human takeover occurs instantaneously. In our actual testing setup, the system runs at 5 Hz, meaning that a full takeover requires 0.2 seconds of human control. The footnoted columns report results under the assumption that each full takeover lasts 1, 2, or 3 seconds.

The results indicate a clear trend: as humans intend to sustain longer takeover periods, the total amount of required intervention decreases. Notably, when the human remains in control for longer durations, more potential takeover events are absorbed within that interval and therefore do not trigger additional interventions.so the takeover frequency are reducing while the human takeover time increase.

\section{Implementation Details}
\label{sec:implementation_details}

Our training pipeline consists of two stages.
For stage 1, we train the {\CmdEncodershort} and the LoRA-adapted LLM to generate action captions conditioned on the drafting and the arrowing prompt. 
The modules are trained on eight A6000 GPUs using a cosine learning rate schedule with a base learning rate of $2\times10^{-5}$, weight decay of 0.05, and a warmup ratio of 0.03, for 5 epochs with a global batch size of 32.
For stage 2, we freeze all other components and train only the action decoder in an end-to-end manner. 
The policy is trained on eight RTX~4090 GPUs using the AdamW optimizer with a learning rate of $1\times10^{-4}$ for 100 epochs and a global batch size of 32. 
The AdamW hyperparameters are set to $\beta_1 = 0.9$ and $\beta_2 = 0.99$.

\end{document}